# Exploiting the Structure: Stochastic Gradient Methods Using Raw Clusters

Zeyuan Allen-Zhu*
Princeton University
zeyuan@csail.mit.edu

Yang Yuan*
Cornell University
yangyuan@cs.cornell.edu

Karthik Sridharan
Cornell University
sridharan@cs.cornell.edu

February 2, 2016†

**Abstract**

The amount of data available in the world is growing faster than our ability to deal with it. However, if we take advantage of the internal *structure*, data may become much smaller for machine learning purposes. In this paper we focus on one of the fundamental machine learning tasks, empirical risk minimization (ERM), and provide faster algorithms with the help from the clustering structure of the data.

We introduce a simple notion of *raw clustering* that can be efficiently computed from the data, and propose two algorithms based on clustering information. Our accelerated algorithm ClusterACDM is built on a novel Haar transformation applied to the dual space of the ERM problem, and our variance-reduction based algorithm ClusterSVRG introduces a new gradient estimator using clustering. Our algorithms outperform their classical counterparts ACDM and SVRG respectively.

## 1 Introduction

For large-scale machine learning applications, $n$, the number of training data examples, is usually very large. To search for the optimal solution, it is often desirable to use *stochastic* gradient methods which only require one (or a batch of) random example(s) from the given training set per iteration in order to form an *estimator* of the true gradient.

For empirical risk minimization problems (ERM) in particular, stochastic gradient methods have received a lot of attention in the past decade. The original stochastic gradient descent (SGD) [4, 29] simply defines the estimator using one random data example and converges slowly. Recently, variance-reduction methods were introduced to improve the running time of SGD [6, 7, 13, 19, 22, 23, 25, 27], and accelerated gradient methods were introduced to further improve the running time when the regularization parameter is small [9, 17, 18, 26, 30].

None of the above cited results, however, have considered the *internal structure* of the dataset, that is, using the stochastic gradient with respect to one data vector $p$ to estimate the stochastic gradients of other data vectors close to $p$. To illustrate why internal structure can be helpful,

*These two authors equally contribute to this paper.

†First version appeared on this date on arXiv. An extended abstract of this paper has appeared in NIPS 2016.

consider the following extreme case: if all the data vectors are located at the same spot, then every stochastic gradient represents the full gradient of the entire dataset. In a non-extreme case, if data vectors form *clusters*, then the stochastic gradient of one data vector could provide a rough estimation for its neighbors. Therefore, one should expect ERM problems to be *easier* if the data vectors are clustered.

More importantly, well-clustered datasets are *abundant* in big-data scenarios. For instance, although there are more than 1 billion of users on Facebook, the intrinsic "feature vectors" of these users can be naturally categorized by the users' occupations, nationalities, etc. As another example, although there are 581,012 vectors in the famous Covtype dataset [8], each representing a 30m x 30m cell in the Roosevelt National Forest of northern Colorado, these vectors can be efficiently categorized into 1,445 clusters of diameter 0.1 — see Section 5. With these examples in mind, we investigate in this paper how to train an ERM problem faster using clustering information.

### 1.1 Known Result and Our Notion of Raw Clustering

In a seminal work by Hofmann et al. published in NIPS 2015 [11], they introduced N-SAGA, the first ERM training algorithm that takes into account the similarities between data vectors. In each iteration, N-SAGA computes the stochastic gradient of one data vector $p$, and uses this information as a biased representative for a small neighborhood of $p$ (say, 20 nearest neighbors of $p$).

In this paper, we focus on a more general and powerful notion of clustering yet capturing only the *minimum* requirement for a cluster to have similar vectors. Assume without loss of generality data vectors have norm at most 1. We say that a partition of the data vectors is an $(s, \delta)$ *raw clustering* if the vectors are divided into $s$ disjoint sets, and the *average* distance between vectors in each set is at most $\delta$. For different values of $\delta$, one can obtain an $(s_\delta, \delta)$ raw clustering where $s_\delta$ is a function on $\delta$. For example, a $(1445, 0.1)$ raw clustering exists for the Covtype dataset that contains 581,012 data vectors. Raw clustering enjoys the following nice properties.

- It allows outliers to exist in a cluster and nearby vectors to be split into multiple clusters.
- It allows large clusters. This is in contrast to N-SAGA which requires each cluster to be very small (say of size 20) due to their algorithmic limitation.

**Computation Overhead.** Since we do not need exactness, raw clusterings can be obtained very efficiently. We directly adopt the approach of Hofmann et al. [11] because finding approximate clustering is the same as finding approximate neighbors. Hofmann et al. [11] proposed to use approximate nearest neighbor algorithms such as LSH [2, 3] and product quantization [10, 12], and we use LSH in our experiments. Without trying hard to optimize the code, we observed that in time $0.3T$ we can detect if good clustering exists, and if so, in time around $3T$ we find the actual clustering. Here $T$ is the running time for a stochastic method such as SAGA to perform $n$ iterations (i.e., one pass) on the dataset.

We repeat three remarks from Hofmann et al. First, the better the clustering quality the better performance we can expect; yet one can always use the trivial clustering as a fallback option. Second, the clustering time should be amortized over multiple runs of the training program: if one performs 30 runs to choose between loss functions and tune parameters, the amortized cost to compute a raw clustering is at most $0.1T$. Third, since stochastic gradient methods are sequential methods, increasing the computational cost in a highly parallelizable way may not affect data throughput.

Note. Clustering can also be obtained for free in some scenarios. If Facebook data are retrieved, one can use the geographic information of the users to form raw clustering. If one works with the CIFAR-10 dataset, the known CIFAR-100 labels can be used as clustering information too [15].

## 1.2 Our New Results

We first observe some limitations of N-SAGA. Firstly, it is *biased* algorithm and does not converge to the objective minimum.[1] Secondly, in order to keep the bias small, N-SAGA only exploits a small neighborhood for every data vector. Thirdly, N-SAGA may need 20 times more computation time per iteration as compared to SAGA or SGD, if 20 is the average neighborhood size. This makes the total running time of N-SAGA in doubt for certain training problems.

We explore in this paper how a given $(s, \delta)$ raw clustering can improve the performance of training ERM problems. We propose two unbiased algorithms that we call ClusterACDM and ClusterSVRG. The two algorithms use different techniques. ClusterACDM uses a novel clustering-based transformation in the dual space, and provides a faster algorithm than ACDM [1, 16] both in practice and in terms of asymptotic worst-case performance. ClusterSVRG is built on top of SVRG [13], but using a new clustering-based gradient estimator to improve the running time.

More specifically, consider *for simplicity* ridge regression where the $\ell_2$ regularizer has weight $\lambda > 0$. The best known non-accelerated methods (such as SAGA [6] and SVRG [6]) and the best known accelerated methods (such as ACDM or AccSDCA [26])) run in time respectively

$$\text{non-accelerated: } \widetilde{O}\Big(nd + \tfrac{d}{\lambda}\Big) \quad \text{and} \quad \text{accelerated: } \widetilde{O}\Big(nd + \tfrac{\sqrt{n}}{\sqrt{\lambda}}d\Big) \tag{1.1}$$

where $d$ is the dimension of the data vectors and the $\widetilde{O}$ notation hides the $\log(1/\varepsilon)$ factor that depends on the accuracy. Accelerated methods converge faster when $\lambda$ is smaller than $1/n$.

Just by looking at the asymptotic worst-case running times in (1.1), we argue that the clustering information can have *drastically* different impacts on the two types of methods.

- In a non-accelerated method, even if all the data vectors were identical so a stochastic gradient would be the same as the full gradient, a non-accelerated method still had to run $\widetilde{O}(\frac{1}{\lambda})$ iterations, yielding a total running time $\widetilde{O}(d/\lambda)$. Therefore, there is little room to improve the asymptotic worst-case running time of a non-accelerated method, even if good clustering information is provided. [2]

- In contrast, for an accelerated method, if all the data vectors were identical, an accelerated method could converge in $\widetilde{O}(\frac{1}{\sqrt{\lambda}})$ iterations, yielding a total running time $O(d/\sqrt{\lambda})$. In other words, one could expect the running time of an accelerated method to be improved from $\widetilde{O}\big(nd + \frac{\sqrt{n}d}{\sqrt{\lambda}}\big)$ to $\widetilde{O}\big(nd + \frac{d}{\sqrt{\lambda}}\big)$ at best.

Our ClusterACDM method is designed to precisely match this suggested performance above. Given an $(s, \delta)$ raw clustering, ClusterACDM enjoys a worst-case running time

Our ClusterACDM method outperforms (1.1) both in terms of theory and practice. Given an $(s, \delta)$ raw clustering, ClusterACDM enjoys a worst-case running time

$$\widetilde{O}\Big(nd + \tfrac{\max\{\sqrt{s}, \sqrt{\delta n}\}}{\sqrt{\lambda}}d\Big) \ . \tag{1.2}$$

In the ideal case when all the feature vectors are identical, ClusterACDM converges in time $\widetilde{O}\big(nd + \frac{d}{\sqrt{\lambda}}\big)$. Otherwise, our running time is asymptotically better than known accelerated methods by a factor $O(\min\{\sqrt{\frac{n}{s}}, \frac{1}{\sqrt{\delta}}\})$ that depends on the clustering quality. [3] Our speed-up also generalizes to other ERM problems as well such as Lasso.

[1]N-SAGA uses the stochastic gradient of one data vector to completely represent its neighbors. This changes the objective value and therefore cannot give very accurate solutions.

[2]However, as we show in Section 4, ClusterSVRG still improves the running time especially at the first a few passes of the dataset.

Our ClusterSVRG matches the best non-accelerated result in (1.1) in the worst-case;[4] however, it enjoys a provably smaller variance than SVRG or SAGA, so runs faster in practice.

**Techniques Behind ClusterACDM.** We highlight our main techniques behind ClusterACDM. Since a cluster of vectors have almost identical directions if $\delta$ is small, we wish to create an auxiliary vector for each cluster representing "moving in the average direction of all vectors in this cluster". Next, we design a stochastic gradient method that, instead of uniformly choosing a random vector, selects those auxiliary vectors with a much higher probability compared with ordinary ones. This could lead to a running time improvement because moving in the direction of an auxiliary vector only costs $O(d)$ running time but exploits the information of the entire cluster.

We implement the above intuition using *optimization* insights. In the dual space of the ERM problem, each variable corresponds to a data example in the primal, and the objective is known to be coordinate-wise smooth with the same smoothness parameter per coordinate. In the preprocessing step, ClusterACDM applies a novel *Haar transformation* on each cluster of the dual coordinates. Haar transformation rotates the dual space, and for each cluster, it automatically reserves a new dual variable that corresponds to the "auxiliary vector" mentioned above. Furthermore, these new dual variables have significantly larger smoothness parameters and therefore will be selected with probability much larger than $1/n$ if one applies a state-of-the-art accelerated coordinate descent method such as ACDM.

### 1.3 To Practitioners

**SVRG vs. SAGA vs. ClusterSVRG.** SVRG becomes a special case of ClusterSVRG when all the data vectors belong to the *same* cluster; SAGA becomes a special case of ClusterSVRG when each data vector belongs to its *own* cluster. We hope that this interpolation helps experimentalists decide between these methods: (1) if the data vectors are pairwise close to each other then use SVRG; (2) if the data vectors are all very separated from each other then use SAGA; and (3) if the data vectors have nice clustering structures (which one can detect using LSH), then use our ClusterSVRG.

**ClusterSVRG vs. ClusterACDM.** Based on both theoretical and experimental results, we have the following suggestions for choosing between ClusterSVRG and ClusterACDM for experimentalists. When the given clusters has large average sizes (thus large cluster diameter), we recommend ClusterACDM and vice versa because ClusterACDM exploits large clusters better. When the regularization parameter $\lambda$ is large, we recommend ClusterSVRG and vice versa. In practice, one may first pick the best $\lambda$ using cross validation, and then decide between ClusterSVRG and ClusterACDM. See Section 5.2 for detailed comparison and reasoning.

**Mini Batch.** Recall that both non-accelerated and accelerated gradient methods have their mini-batch variants, such as mini-batch SVRG [14], mini-batch AccSDCA [24], and mini-batch SPDC [30]. Instead of computing a stochastic gradient with respect to a single training data example, these algorithms perform updates with respect to a random subset of the data examples, known as a *mini batch*. These results are therefore orthogonal to the present work, and adopting mini-batch updates can further improve performance for a clustering-based method. For instance, by selecting a random data example from each cluster as a mini batch in each iteration, one can further reduce the variance of the gradient estimator in ClusterSVRG and improve its convergence.

[3] Since $s$ increases as $\delta$ decreases, this speed up factor is maximized when $s/\delta = n$ in theory. In practice, however, the performance of ClusterACDM is not very sensitive to the quality of clusters. As we shown in Section 5.3, as long as the given clustering is not degenerated, we obtain similarly fast convergence rates.

[4] The asymptotic worst-case running time for non-accelerated methods in (1.1) cannot be improved in general, even if a perfect clustering (i.e., $\delta = 0$) is given.

### 1.4 Other Related Work

Reddi et al. [21] also proposes a framework unifying SVRG and SAGA. ClusterSVRG is different from theirs as we compare SVRG and SAGA from the perspective of data clustering, while their framework generalizes several algorithms using a universal ScheduleUpdate subroutine. With the help from this clustering perspective, a more fundamental difference between SAGA and SVRG is revealed and verified by experiments in this paper: SAGA implicitly treats each data vector as a separate cluster, while SVRG implicitly treats all the vectors as a single cluster.

Researchers also design accelerated stochastic methods via a black-box reduction to non-accelerated ones [9, 17]. The worst-case running times of these methods match that of AccSDCA. However, we can not use these reductions to directly get ClusterACDM from ClusterSVRG, since in ClusterACDM we use different techniques such as the Haar transformation and non-uniform sampling, which do not appear in either ClusterSVRG or the reduction frameworks.

**Other Related Work.** ClusterACDM can be viewed as "preconditioning" the data matrix from the dual variable side. Recently, preconditioning received some attentions in machine learning. In particular, non-uniform sampling can be viewed as using diagonal preconditioners [1, 31]. However, diagonal preconditioning has nothing to do with clustering: for instance, if all data vectors have the same Euclidean norm, the cited results are identical to SVRG or APCG so do not exploit the clustering information. Some authors also study preconditioning from the primal side using SVD [28]. This is different from us because for instance when all the data vectors are same (thus forming a perfect cluster), the cited result reduces to SVRG and does not improve the running time. Furthermore, even if one performs SVD on the dual side to exploit clustering information, the time needed to do so is much larger than computing raw clustering.

**Roadmap.** We introduce necessary notations and concepts in Section 2. In Section 3 and 4, we respective describe our algorithms ClusterACDM and ClusterSVRG and discuss the high level intuitions behind the proofs, while deferring the detailed analysis in Appendix A and Appendix B respectively. In Section 5 we present our experimental results.

## 2 Preliminaries

### 2.1 Raw Clustering

Given a dataset consisting of $n$ vectors $\{a_1, \dots, a_n\} \subset \mathbb{R}^d$, we assume without loss of generality that $\|a_i\|_2 \le 1$ for each $i \in [n]$. Let a *clustering* of the dataset be a partition of the indices $[n] = S_1 \cup \cdots \cup S_s$. We call each set $S_c$ a *cluster* and use $n_c = |S_c|$ to denote its size. It satisfies $\sum_{c=1}^{s} n_c = n$. We are interested in the following quantification that estimates the clustering quality:

**Definition 2.1** (raw clustering on vectors)**.** *We say a partition $[n] = S_1 \cup \cdots \cup S_s$ is an $(s, \delta)$ raw clustering for the vectors $\{a_1, \dots, a_n\}$ if, on average, the pairwise distances between vectors in each cluster $S_c$ are at most $\delta$; or more precisely, for every cluster $S_c$ it satisfies*

$$\frac{1}{|S_c|^2} \sum_{i,j \in S_c} \|a_i - a_j\|^2 \le \delta \ .$$

We call it a *raw* clustering because the above definition captures the *minimum* requirement for each cluster to have similar vectors. For instance, the above "average" definition allows a few outliers to exist in each cluster and allows nearby vectors to be split into different clusters.

Raw clustering of the dataset is very easy to obtain: we include in Section 5.1 a simple and efficient algorithm for computing an $(s_\delta, \delta)$ raw clustering of any quality $\delta$. A similar assumption like our $(s, \sigma)$ raw clustering assumption in Definition 2.1 was also introduced by Hofmann et al. [11].

### 2.2 Convex Optimization

Recall some classical definitions on the strong convexity and the smoothness of a function.

**Definition 2.2** (Smoothness and strong convexity). *For a convex function $g \colon \mathbb{R}^n \to \mathbb{R}$,*

- *$g$ is $\sigma$-strongly convex if $\forall x, y \in \mathbb{R}^n$, it satisfies $g(y) \geq g(x) + \langle \nabla g(x), y - x\rangle + \frac{\sigma}{2}\|x - y\|^2$.*
- *$g$ is $L$-smooth if $\forall x, y \in \mathbb{R}^n$, it satisfies $\|\nabla g(x) - \nabla g(y)\| \leq L\|x - y\|$.*
- *$g$ is coordinate-wise smooth with parameters $(L_1, L_2, \ldots, L_n)$, if*

$$\forall x \in \mathbb{R}^n,\ \forall \delta > 0,\ \forall i \in [n] \colon |\nabla_i g(x + \delta \mathbf{e}_i) - \nabla_i g(x)| \leq L_i \cdot \delta\ \ .$$

*If $g$ is twice differentiable, this requirement can be simplified as $\nabla^2_{ii} g(x) \leq L_i$ for all $x \in \mathbb{R}^n$.*

For strongly convex and coordinate-wise smooth functions $g$, one can apply the accelerated coordinate descent algorithm (ACDM) to minimize $g$:

**Theorem 2.3** (ACDM). *If $g(x)$ is $\sigma$-strongly convex and coordinate-wise smooth with parameters $(L_1, \ldots, L_n)$, the non-uniform accelerated coordinate descent method of [1] produces an output $y$ satisfying $g(y) - \min_x g(x) \leq \varepsilon$ in*

$$O\Big(\sum_i \sqrt{L_i/\sigma} \cdot \log(1/\varepsilon)\Big)$$

*iterations. Each iteration runs in time proportional to the computation of a coordinate gradient $\nabla_i g(\cdot)$ of $g$.*

**Remark 2.4.** Accelerated coordinate descent admits several variants such as APCG [18], ACDM [16], and NU_ACDM [1]. These variants agree on the running time when $L_1 = \cdots = L_n$, but NU_ACDM is the fastest when $L_1, \ldots, L_n$ are *non-uniform*. More specifically, NU_ACDM selects a coordinate $i$ with probability proportional to $\sqrt{L_i}$. In contrast, ACDM samples coordinate $i$ with probability proportional to $L_i$ and APCG samples $i$ with probability $1/n$. We refer to NU_ACDM as the accelerated coordinate descent method (ACDM) in this paper.

## 3 ClusterACDM Algorithm

Our ClusterACDM method is an *accelerated* stochastic gradient method just like AccSDCA [26], APCG [18], ACDM [1, 16], SPDC [30], etc. Consider a regularized least-square problem

$$\text{Primal:} \quad \min_{x \in \mathbb{R}^d} \Big\{ P(x) \stackrel{\text{def}}{=} \frac{1}{2n} \sum_{i=1}^n (\langle a_i, x\rangle - l_i)^2 + r(x) \Big\}\ \ , \tag{3.1}$$

where each $a_i \in \mathbb{R}^d$ is the feature vector of a training example and $l_i$ is the label of $a_i$. Problem (3.1) becomes *ridge regression* when $r(x) = \frac{\lambda}{2}\|x\|_2^2$, and becomes *Lasso* when $r(x) = \lambda\|x\|_1$. One of the state-of-the-art *accelerated* stochastic gradient methods to solve (3.1) is through its dual.

**Algorithm 1** ClusterACDM

**Input:** a raw clustering $S_1 \cup \cdots \cup S_s$.

1: Apply cluster-based Haar transformation $H_{\mathsf{cl}}$ to get the transformed objective $D'(y')$.
2: Run ACDM to minimize $D'(y')$
3: Transform the solution of $D'(y')$ back to the original space.

Consider the following equivalent dual formulation of (3.1) (see for instance [18] for the detailed proof):

$$\text{Dual:} \quad \min_{y \in \mathbb{R}^n} \Big\{ D(y) \stackrel{\text{def}}{=} \frac{1}{2n}\|y\|^2 + \frac{1}{n}\langle y, l \rangle + r^*\big(-\frac{1}{n}Ay\big) = \frac{1}{n}\sum_{i=1}^{n}\big(\frac{1}{2}y_i^2 + y_i \cdot l_i\big) + r^*\Big(-\frac{1}{n}\sum_{i=1}^{n} y_i a_i\Big)\Big\} \ , \tag{3.2}$$

where $A = [a_1, a_2, \ldots, a_n] \in \mathbb{R}^{d \times n}$ and $r^*(y) \stackrel{\text{def}}{=} \max_w y^T w - r(w)$ is the Fenchel dual of $r(w)$.

## 3.1 Previous Solutions

If $r(x)$ is $\lambda$-strongly convex in $P(x)$, the dual objective $D(y)$ is both strongly convex and smooth. The following lemma is due to [18] but is also proved in our appendix for completeness.

**Lemma 3.1.** *If $r(x)$ is $\lambda$-strongly convex, then $D(y)$ is $\sigma = \frac{1}{n}$ strongly convex and coordinate-wise smooth with parameters $(L_1, \ldots, L_n)$ for $L_i = \frac{1}{n} + \frac{1}{\lambda n^2}\|a_i\|^2$.*

For this reason, the authors of [18] proposed to apply accelerated coordinate descent (such as their APCG method) to minimize $D(y)$.[5] Assuming without loss of generality $\|a_i\|^2 \le 1$ for $i \in [n]$, we have $L_i \le \frac{1}{n} + \frac{1}{\lambda n^2}$. Using Theorem 2.3 on $D(\cdot)$, we know that ACDM produces an $\varepsilon$-approximate dual minimizer $y$ in

$$O\Big(\sum_i \sqrt{L_i/\sigma}\log(1/\varepsilon)\Big) = O\Big(n\sqrt{\frac{\frac{1}{n}+\frac{1}{\lambda n^2}}{1/n}}\log(1/\varepsilon)\Big) = \widetilde{O}(n + \sqrt{n/\lambda})$$

iterations, and each iteration runs in time proportional to the computation of $\nabla_i D(y)$ which is $O(d)$. This total running time $\widetilde{O}(nd + \sqrt{n/\lambda} \cdot d)$ is the fastest for solving (3.1) when $r(x)$ is $\lambda$-strongly convex.

Due to space limitation, in the main body we *only focus* on the case when $r(x)$ is strongly convex; the non-strongly convex case (such as Lasso) can be reduced to this case. See Remark A.1 in appendix.

## 3.2 Our New Algorithm

Each dual coordinate $y_i$ naturally corresponds to the $i$-th feature vector $a_i$. Therefore, given a raw clustering $[n] = S_1 \cup S_2 \cup \cdots \cup S_s$ of the dataset, we can partition the coordinates of the dual vector $y \in \mathbb{R}^n$ into $s$ blocks each corresponding to a cluster. Without loss of generality, we assume the coordinates of $y$ are sorted in the order of the cluster indices. In other words, we write $y = (y_{S_1}, \ldots, y_{S_s})$ where each $y_{S_c} \in \mathbb{R}^{n_c}$.

ClusterACDM transforms the dual objective (3.2) into an equivalent form, by performing an $n_c$-dimensional Haar transformation on the $c$-th block of coordinates for every $c \in [s]$. Formally,[6]

[5] They showed that defining $x = \nabla r^*(-Ay/n)$, if $y$ is a good approximate minimizer of the dual objective $D(y)$, $x$ is also a good approximate minimizer of the primal objective $P(x)$.

[6] We note that an $n$-dimensional Haar transformation is often defined in the literature only for $n$ being an integral power of 2; in contrast, we generalize them into arbitrary integral dimensions for our application in this paper.

**Definition 3.2.** *Let* $R_2 \stackrel{\text{def}}{=} \begin{bmatrix} 1/\sqrt{2} & -1/\sqrt{2} \end{bmatrix}$, $R_3 \stackrel{\text{def}}{=} \begin{bmatrix} \sqrt{2}/\sqrt{3} & -\sqrt{2}/(2\sqrt{3}) & -\sqrt{2}/(2\sqrt{3}) \\ 0 & 1/\sqrt{2} & -1/\sqrt{2} \end{bmatrix}$, *and more generally*

$$R_n \stackrel{\text{def}}{=} \begin{bmatrix} \frac{1/a}{\sqrt{1/a+1/b}} \cdots \frac{1/a}{\sqrt{1/a+1/b}} & \frac{-1/b}{\sqrt{1/a+1/b}} \cdots \frac{-1/b}{\sqrt{1/a+1/b}} \\ R_a & 0 \\ 0 & R_b \end{bmatrix} \in \mathbb{R}^{(n-1)\times n}$$

*for* $a = \lfloor n/2 \rfloor$ *and* $b = \lceil n/2 \rceil$. *Then, define the* $n$*-dimensional* (normalized) Haar matrix *as*

$$H_n \stackrel{\text{def}}{=} \begin{bmatrix} 1/\sqrt{n} \cdots 1/\sqrt{n} \\ R_n \end{bmatrix} \in \mathbb{R}^{n\times n}$$

We give a few examples of Haar matrices in Example A.2 in Appendix A. It is easy to verify that

**Lemma 3.3.** *For every* $n$, $H_n^T H_n = H_n H_n^T = I$, *so* $H_n$ *is a unitary matrix.*

**Definition 3.4.** *Given a clustering* $[n] = S_1 \cup \cdots \cup S_s$, *define the following* cluster-based Haar transformation $H_{\mathsf{cl}} \in \mathbb{R}^{n\times n}$ *that is a block diagonal matrix:*

$$H_{\mathsf{cl}} \stackrel{\text{def}}{=} \begin{pmatrix} H_{|S_1|} & 0 & 0\cdots 0 & 0 \\ 0 & H_{|S_2|} & 0\cdots 0 & 0 \\ \vdots & \vdots & \ddots & \vdots \\ 0 & 0 & 0\cdots 0 & H_{|S_s|} \end{pmatrix} .$$

*Accordingly, we apply the unitary transformation* $H_{\mathsf{cl}}$ *on (3.2) and consider*

$$\min_{y'\in\mathbb{R}^n} \Big\{ D'(y') \stackrel{\text{def}}{=} \frac{1}{2n}\|y'\|^2 + \frac{1}{n}\langle y', H_{\mathsf{cl}} l\rangle + r^*\big(-\frac{1}{n} A H_{\mathsf{cl}}^T y'\big) \Big\} . \tag{3.3}$$

*We call* $D'(y')$ *the* transformed objective *function.*

It is clear that the minimization problem (3.3) is equivalent to (3.2) by transforming $y = H_{\mathsf{cl}}^T y'$. Now, our ClusterACDM algorithm applies ACDM on minimizing this transformed objective $D'(y')$.

We claim the following running time of ClusterACDM and discuss the high-level intuition in the main body. We defer detailed analysis to Appendix A.

**Theorem 3.5.** *If* $r(\cdot)$ *is* $\lambda$*-strongly convex and an* $(s, \delta)$ *raw clustering is given, then ClusterACDM outputs an* $\varepsilon$*-approximate minimizer of* $D(\cdot)$ *in time* $T = O\big(nd + \frac{\max\{\sqrt{s},\sqrt{\delta n}\}}{\sqrt{\lambda}} d\big)$ .

Comparing to the complexity of APCG, ACDM, or AccSDCA (see (1.1)), ClusterACDM is faster by a factor that is up to $\Omega\big(\min\{\sqrt{n/s}, \sqrt{1/\delta}\}\big)$.

**High-Level Intuition.** To see why Haar transformation is helpful, we focus on one cluster $c \in [s]$. Assume without loss of generality that cluster $c$ has vectors $a_1, a_2, \cdots, a_{n_c}$. After applying Haar transformation, the new columns $1, 2, \ldots, n_c$ of matrix $AH_{\mathsf{cl}}^T$ become weighted combinations of $a_1, a_2, \cdots, a_{n_c}$, and the weights are determined by the entries in the corresponding rows of $H_{n_c}$.

Observe that every row except the first one in $H_{n_c}$ has its entries sum up to 0. Therefore, columns $2, \ldots, n_c$ in $AH_{\mathsf{cl}}^T$ will be close to zero vectors and have small norms. In contrast, since the first row in $H_{n_c}$ has all its entries equal to $1/\sqrt{n_c}$, the first column of $AH_{\mathsf{cl}}^T$ becomes $\sqrt{n_c}\cdot\frac{a_1+\cdots+a_{n_c}}{n_c}$,

the scaled average of all vectors in this cluster. It has a large Euclidean norm. See the following illustration:

$$\big[a_1,\ a_2,\ \dots, a_{n_c}\big] H_{n_c} = \Big[\sqrt{n_c}\cdot\frac{a_1+\cdots+a_{n_c}}{n_c},\ \ \approx 0, \cdots, \approx 0\Big]$$

The first column after Haar transformation can be viewed as an auxiliary feature vector representing the entire cluster. If we run ACDM with respect to this new matrix, and whenever this auxiliary column is selected, it represents "moving in the average direction of all vectors in this cluster". Since this single auxiliary column cannot represent the entire cluster, the remaining $n_c-1$ columns serve as helpers that ensure that the algorithm is *unbiased* (i.e., converges to the exact minimizer).

Most importantly, as discussed in Remark 2.4, ACDM is a stochastic method that samples a dual coordinate $i$ (thus a primal feature vector $a_i$) with a probability proportional to its square-root coordinate-smoothness (thus roughly proportional to $\|a_i\|$). Since auxiliary vectors have much larger Euclidean norms, we expect them to be sampled with probabilities much larger $1/n$. This is how the faster running time is obtained in Theorem 3.5.

Remark. The speed-up of ClusterACDM depends on how much "non-uniformity" the underlying coordinate descent method can utilize. Therefore, no speed-up can be obtained if one applies APCG instead of the NU_ACDM which is optimally designed to utilize coordinate non-uniformity. Similarly, if one uses ACDM instead of NU_ACDM, the speed-up becomes square root of what we have.

# 4 ClusterSVRG Algorithm

Our ClusterSVRG is a *non-accelerated* stochastic gradient method just like SVRG [13], SAGA [6], SDCA [25], etc. It directly works on minimizing the primal objective (similar to SVRG and SAGA):

$$\min_{x\in\mathbb{R}^d}\Big\{F(x)\stackrel{\text{def}}{=}f(x)+\Psi(x)\stackrel{\text{def}}{=}\frac{1}{n}\sum_{i=1}^{n}f_i(x)+\Psi(x)\Big\}\ . \tag{4.1}$$

Here, $f(x)=\frac{1}{n}\sum_{i=1}^n f_i(x)$ is the finite average of $n$ functions, each $f_i(x)$ is convex and $L$-smooth, and $\Psi(x)$ is a simple (but possibly non-differentiable) convex function, sometimes called the proximal function. We denote $x^*$ as a minimizer of (4.1).

Recall that stochastic gradient methods work as follows. At every iteration $t$, they perform updates $x^t \leftarrow x^{t-1}-\eta\widetilde{\nabla}^{t-1}$ for some step length $\eta>0$,[7] where $\widetilde{\nabla}^{t-1}$ is the so-called *gradient estimator* and its expectation had better equal the full gradient $\nabla f(x^{t-1})$. It is a known fact that the faster the variance $\mathbf{Var}[\widetilde{\nabla}^{t-1}]$ diminishes, the faster the underlying method converges. [13]

For instance, SVRG defines the estimator as follows. It has an outer loop of epochs. At the beginning of each epoch, SVRG records the current iterate $x$ as a snapshot point $\widetilde{x}$, and computes its full gradient $\nabla f(\widetilde{x})$. In each inner iteration within an epoch, SVRG defines $\widetilde{\nabla}^{t-1}\stackrel{\text{def}}{=}\frac{1}{n}\sum_{j=1}^n \nabla f_j(\widetilde{x})+\nabla f_i(x^{t-1})-\nabla f_i(\widetilde{x})$ where $i$ is a random index in $[n]$. SVRG usually chooses the epoch length $m$ to be $2n$, and it is known that $\mathbf{Var}[\widetilde{\nabla}^{t-1}]$ approaches to zero as $t$ increases. We denote by $\widetilde{\nabla}^{t-1}_{\text{SVRG}}$ this choice of $\widetilde{\nabla}^{t-1}$ for SVRG.

In ClusterSVRG, we define the gradient estimator $\widetilde{\nabla}^{t-1}$ based on clustering information. Given a clustering $[n]=S_1\cup\cdots\cup S_s$ and denoting by $c_i\in[s]$ the cluster that index $i$ belongs to, we define

$$\widetilde{\nabla}^{t-1}\stackrel{\text{def}}{=}\frac{1}{n}\sum_{j=1}^{n}\big(\nabla f_j(\widetilde{x})+\zeta_{c_j}\big)+\nabla f_i(x^{t-1})-\big(\nabla f_i(\widetilde{x})+\zeta_{c_i}\big)\ .$$

[7] Or more generally the proximal updates $x^t\leftarrow \arg\min_x\big\{\frac{1}{2\eta}\|x-x^{t-1}\|^2+\langle\widetilde{\nabla}^{t-1},x\rangle+\Psi(x)\big\}$ if $\Psi(x)$ is nonzero.

**Algorithm 2** ClusterSVRG

**Input:** Epoch length $m$ and learning rate $\eta$, a raw clustering $S_1 \cup \cdots \cup S_s$.

$x^0, \overline{x} \leftarrow$ initial point, $t \leftarrow 0$.
**for** epoch $\leftarrow 0$ **to** MaxEpoch **do**
  $\widetilde{x} \leftarrow x^t$, and $(\zeta_1, \ldots, \zeta_s) \leftarrow (0, \ldots, 0)$
  **for** iter $\leftarrow 1$ **to** $m$ **do**
    $t \leftarrow t+1$ and choose $i$ uniformly at random from $\{1, \cdots, n\}$
    $x^t \leftarrow x^{t-1} - \eta \left( \frac{1}{n} \sum_{j=1}^{n} \left( \nabla f_j(\widetilde{x}) + \zeta_{c_j} \right) + \nabla f_i(x^{t-1}) - \left( \nabla f_i(\widetilde{x}) + \zeta_{c_i} \right) \right)$
    **Option I:** $\zeta_{c_i} \leftarrow \nabla f_i(x^{t-1}) - \nabla f_i(\widetilde{x})$
    **Option II: if** *iter* **mod** $s = 0$ **then** for all $c = 1, \ldots, s$,
      $\zeta_c \leftarrow \nabla f_j(x^{t-1}) - \nabla f_j(\widetilde{x})$ where $j$ is randomly chosen from $S_c$.
  **end for**
**end for**

Above, for each cluster $c$ we introduce an additional $\zeta_c$ term that can be defined in one of the following two ways. Initializing $\zeta_c = 0$ at the beginning of the epoch for each cluster $c$. Then,

- In Option I, after each iteration $t$ is completed and suppose $i$ is the random index chosen at iteration $t$, we update $\zeta_{c_i} \leftarrow \nabla f_i(x^{t-1}) - \nabla f_i(\widetilde{x})$.
- In Option II, we divide an epoch into subepochs of length $s$ each (recall $s$ is the number of clusters). At the beginning of each subepoch, for each cluster $c \in [s]$, we define $\zeta_c \leftarrow \nabla f_j(\overline{x}) - \nabla f_j(\widetilde{x})$. Here, $\overline{x}$ is the last iterate of the previous subepoch and $j$ is a random index in $S_c$.

We summarize both options in Algorithm 2. Note that Option I gives simpler intuition but Option II leads to a simpler proof. Notice that our gradient estimator is unbiased: $\mathbb{E}_i[\widetilde{\nabla}^{t-1}] = \nabla f(x^{t-1})$.

The intuition behind our new choice of $\widetilde{\nabla}^{t-1}$ can be understood as follows. Observe that in the SVRG estimator $\widetilde{\nabla}^{t-1}_{\mathrm{SVRG}}$, each term $\nabla f_j(\widetilde{x})$ can be viewed as a *"guess term"* of the true gradient $\nabla f_j(x^{t-1})$ for function $f_j$. However, these guess terms may be very "outdated" because $\widetilde{x}$ can be $m = 2n$ iterations away from $x^{t-1}$, and therefore contribute to a large variance.

We use raw clusterings to improve these guess terms and reduce the variance. If function $f_j$ belongs to cluster $c$, then our Option I uses $\nabla f_j(\widetilde{x}) + \nabla f_k(x^t) - \nabla f_k(\widetilde{x})$ as the *new* guess of $\nabla f_j(x^t)$, where $t$ is the last time cluster $c$ was accessed and $k$ is the index of the vector in this cluster that was accessed. This new guess only has an "outdatedness" of roughly $s$ that could be much smaller than $n$.

Due to space limitation, we defer all technical details of ClusterSVRG to Appendix B and B.3.

# 5 Experiments

We conduct experiments for three datasets that can be found on the LibSVM website [8]: Covtype.binary, SensIT (combined scale), and News20.binary. To make easier comparison across datasets, we scale every vector by the average Euclidean norm of all the vectors. This step is for comparison only and not necessary in practice. Note that Covtype and SensIT are two datasets where the feature vectors have a nice clustering structure; in contrast, dataset News20 cannot be well clustered and we include it for comparison purpose only.

## 5.1 Clustering and Haar Transformation

We use the approximate nearest neighbor algorithm library E2LSH [2] to compute raw clusterings. Since this is not the main focus of our paper, we include our implementation in Appendix D. The running time needed for raw clustering is reasonable. In Table 1 in the appendix, we list the running time (1) to sub-sample and detect if good clustering exists and (2) to compute the actual clustering. We also list the one-pass running time of SAGA using sparse implementation for comparison.

We conclude two things from Table 1. First, in about the same time as SAGA performing 0.3 pass on the datasets, we can detect clustering structure in the dataset for a given diameter $\delta$. This is a fast-enough preprocessing step to help experimentalists choose to use clustering-based methods or not. Second, in about the same time as SAGA performing 3 passes on well-clustered datasets such as Covtype and SensIT, we obtain the actual raw clustering. As emphasized in the introduction, we view the time needed for clustering as *negligible*. This not only because 0.3 and 3 are small quantities as compared to the average number of passes needed to converge (which is usually around 20). It is also because the clustering time is usually amortized over multiple runs of the training algorithm due to different data analysis tasks, parameter tunings, etc.

In ClusterACDM, we need to pre-compute matrix $AH_{\mathsf{cl}}^T$ using Haar transformation. This can be efficiently implemented thanks to the sparsity of Haar matrices. In Table 2 in the appendix, we see that the time needed to do so is roughly 2 passes of the dataset. Again, this time should be amortized over multiple runs of the algorithm so is negligible.

## 5.2 Performance Comparison

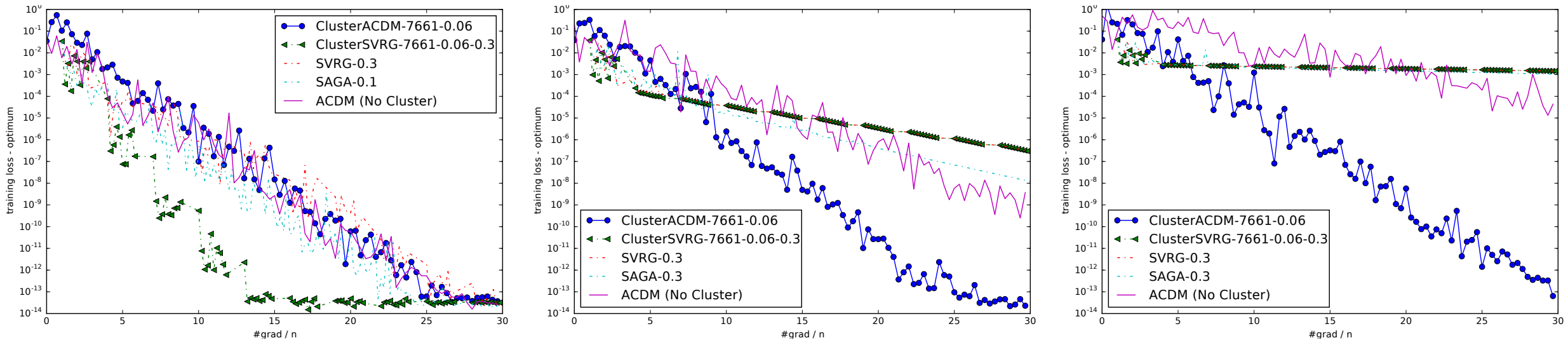

(a) Covtype, Ridge $\lambda = 10^{-5}$ (b) Covtype, Ridge $\lambda = 10^{-6}$ (c) Covtype, Ridge $\lambda = 10^{-7}$

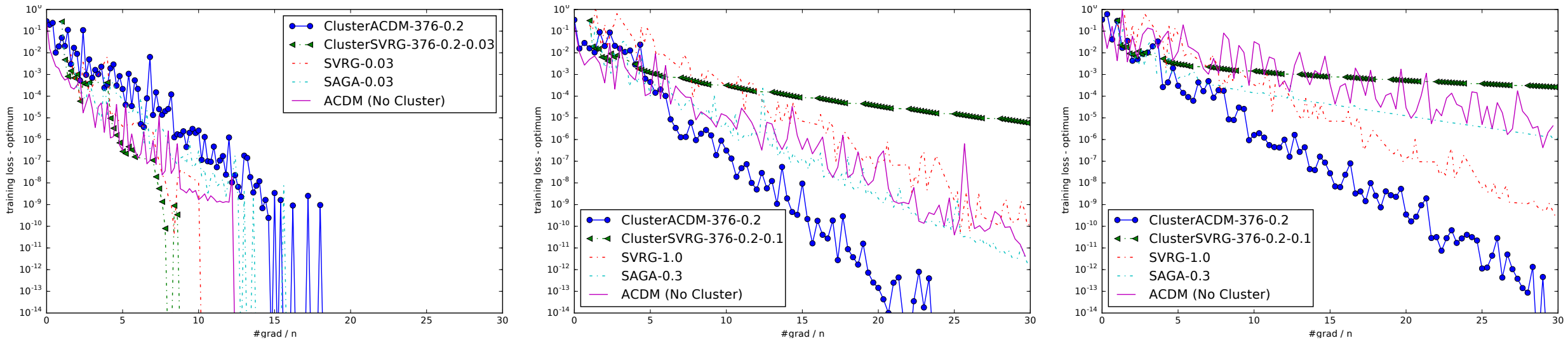

(d) SensIT, Ridge $\lambda = 10^{-3}$ (e) SensIT, Ridge $\lambda = 10^{-5}$ (f) SensIT, Ridge $\lambda = 10^{-6}$

Figure 1: Selected plots on ridge regression. For Lasso and more detailed comparisons, see Appendix

We compare our algorithms with SVRG, SAGA and ACDM. We use default epoch length $m = 2n$ and Option I for SVRG. We use $m = 2n$ and Option I for ClusterSVRG. We consider ridge and Lasso regressions, and denote by $\lambda$ the weight of the $\ell_2$ regularizer for ridge or the $\ell_1$ regularizer for Lasso.

**Parameters.** For SVRG and SAGA, we tune the best step size for each test case. [8] To make our comparison even stronger, instead of tuning the best step size for ClusterSVRG, we simply set it to be either the best of SVRG or the best of SAGA in each test case. For ACDM and ClusterACDM, the step size is computed automatically so tuning is unnecessary.

For Lasso, because the objective is not strongly convex, one has to add a dummy $\ell_2$ regularizer on the objective in order to run ACDM or ClusterACDM. (This step is needed for every accelerated method including AccSDCA, APCG or SPDC.) We choose this dummy regularizer to have weight $10^{-7}$ for Covtype and SenseIT, and weight $10^{-6}$ for News20.[9]

**Plot Format.** In our plots, the $y$-axis represents the objective distance to the minimizer, and the $x$-axis represents the number of passes of the dataset. (The snapshot computation of SVRG and ClusterSVRG counts as one pass.) In the legend, we use the format

- "ClusterSVRG–$s$–$\delta$–stepsize" for ClusterSVRG,
- "ClusterACDM–$s$–$\delta$" for ClusterACDM.
- "SVRG/SAGA–stepsize" for SVRG or SAGA.
- "ACDM (no Cluster)" for the vanilla ACDM without using any clustering info.[10]

**Results.** Our comprehensive experimental plots are included only in the appendix, see Figure 2, 3, 4, 5), 6, and 7. Due to space limitation, here we simply compare all the algorithms on ridge regression for datasets SensIT and Covtype by choosing only one representative clustering, see Figure 1.

Generally, ClusterSVRG outperforms SAGA/SVRG when the regularizing parameter $\lambda$ is *large.* [11] ClusterACDM outperforms all other algorithms when $\lambda$ is *small.* This is because accelerated methods outperform non-accelerated ones with smaller values of $\lambda$, and the complexity of ClusterACDM outperforms ACDM more when $\lambda$ is smaller (compare (1.1) with (1.2)).[12]

Our other findings can be summarized as follows. Firstly, dataset News20 does not have a nice clustering structure but our ClusterSVRG and ClusterACDM still perform comparably well to SVRG and ACDM respectively. Secondly, the performance of ClusterSVRG is slightly better with clustering that has smaller diameter $\delta$. In contrast, ClusterACDM with larger $\delta$ performs slightly better. This is because ClusterACDM can take advantage of very large but low-quality clusters, and this is a very appealing feature in practice. Thirdly, SensIT is a dataset where all feature vectors are close in space. In this case, SVRG performs well because it implicitly treats all the vectors as a single cluster, as discussed in Section 4.

## 5.3 Sensitivity on Clustering

In Figure 8 in appendix, we plot the performance curves of ClusterSVRG and ClusterACDM for SensIT and Covtype, with 7 different clusterings. From the plots we claim that ClusterSVRG and ClusterACDM are very insensitive to the clustering quality. As long as one does not choose the most extreme clustering, the performance improvement due to clustering can be significant. Moreover, ClusterSVRG is slightly faster if the clustering has relatively smaller diameter $\delta$ (say, below 0.1), while the ClusterACDM can be fast even for very large $\delta$ (say, around 0.6).

[8] We tune the best step size in the set $\{3 \times 10^k, 10^k : k \in \mathbb{Z}\}$ for each dataset, for each training problem, and for each $\lambda$.

[9] Choosing large dummy regularizer makes the algorithm converge faster but to a worse minimum, and vice versa. In our experiments, we find these choices reasonable for our datasets. Since our main focus is to compare ClusterACDM with ACDM, as long as we choose the same dummy regularizer our the comparison is *fair*.

[10] ACDM has slightly better performance compared to APCG, so we adopt ACDM in our experiments [1]. Moreover, ACDM supports the non-uniform sampling that is needed in order to obtain the result of ClusterACDM. Furthermore, our comparison is fair because ClusterACDM and ACDM are implemented in the same manner.

[11] This is because for a better regularized objective, the updates between consecutive iterates are smaller. Therefore, the $\xi$ parameter in Assumption B.3 is smaller, and (UB1) dominates (UB2) for a longer time in Lemma B.5.

[12] The best choice of $\lambda$ usually requires cross-validation. For instance, by performing a 10-fold cross validation, one can figure out that the best $\lambda$ is around $10^{-6}$ for SensIT Ridge, $10^{-5}$ for SensIT Lasso, $10^{-7}$ for Covtype Ridge, and $10^{-6}$ for Covtype Lasso. Therefore, for these two datasets ClusterACDM is preferred.

## Acknowledgement

We thank Jinyang Gao for fruitful discussions.

# Appendix

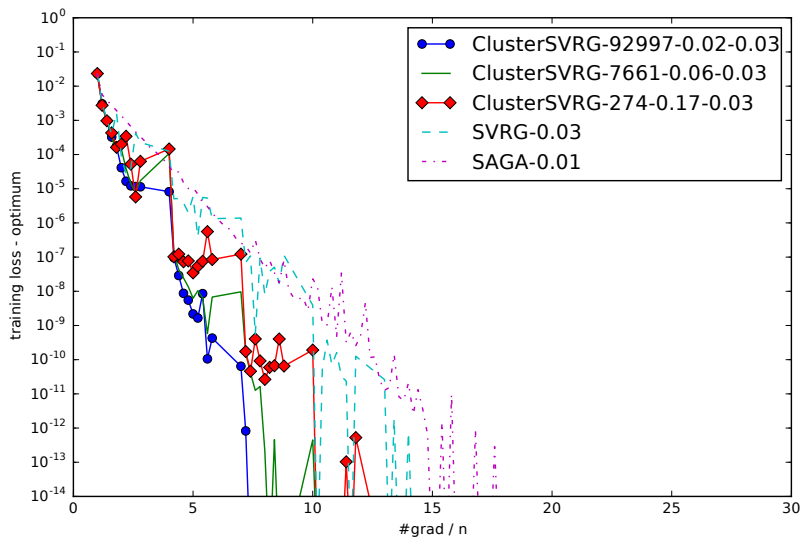

(a) Covtype, Ridge $\lambda = 10^{-4}$

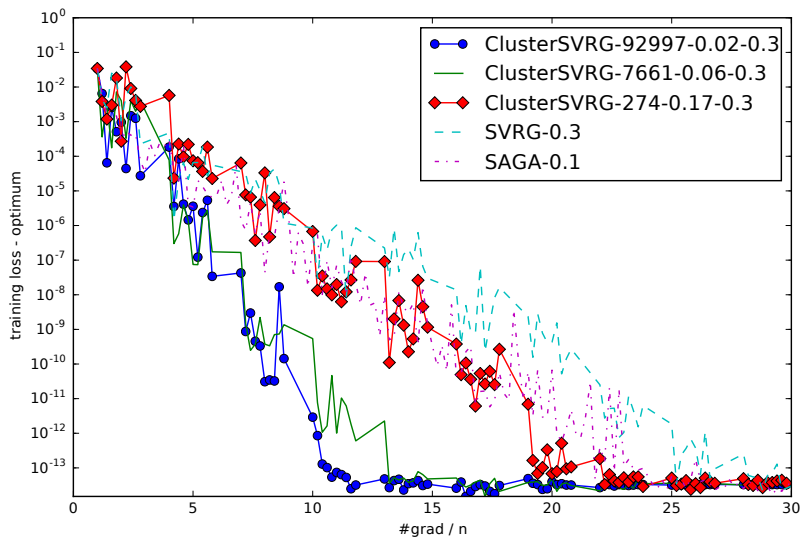

(b) Covtype, Ridge $\lambda = 10^{-5}$

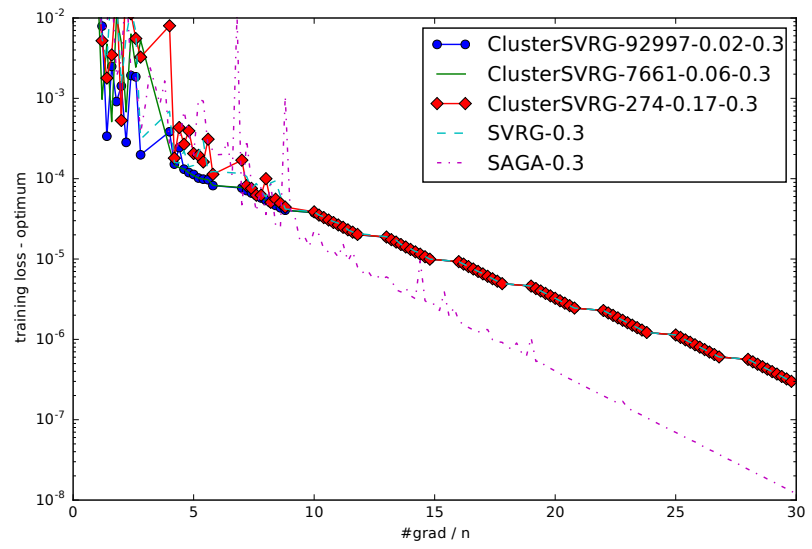

(c) Covtype, Ridge $\lambda = 10^{-6}$

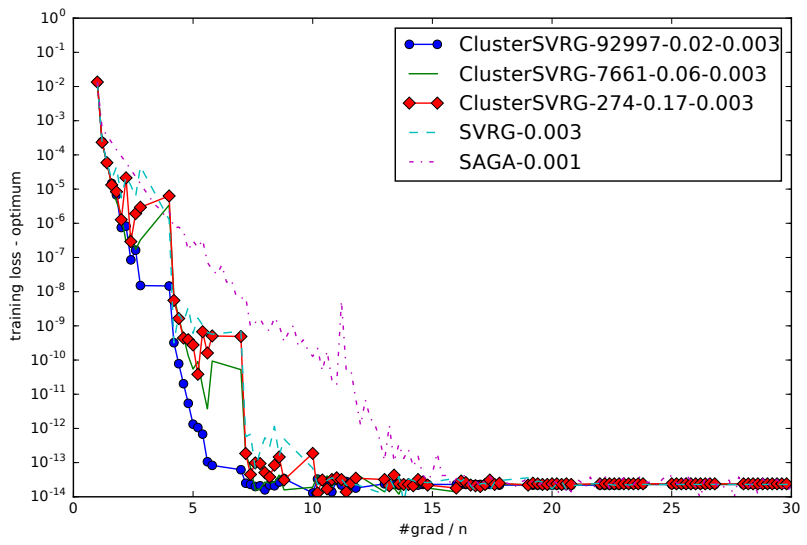

(d) Covtype, Lasso $\lambda = 10^{-3}$

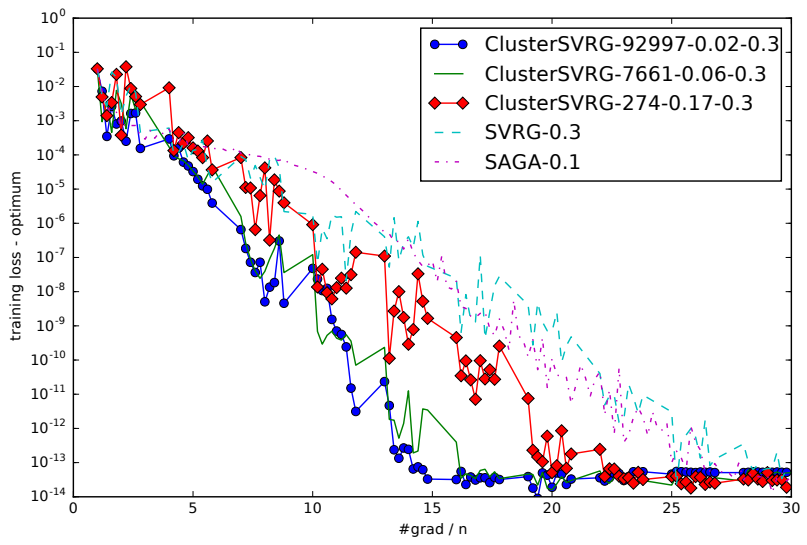

(e) Covtype, Lasso $\lambda = 10^{-4}$

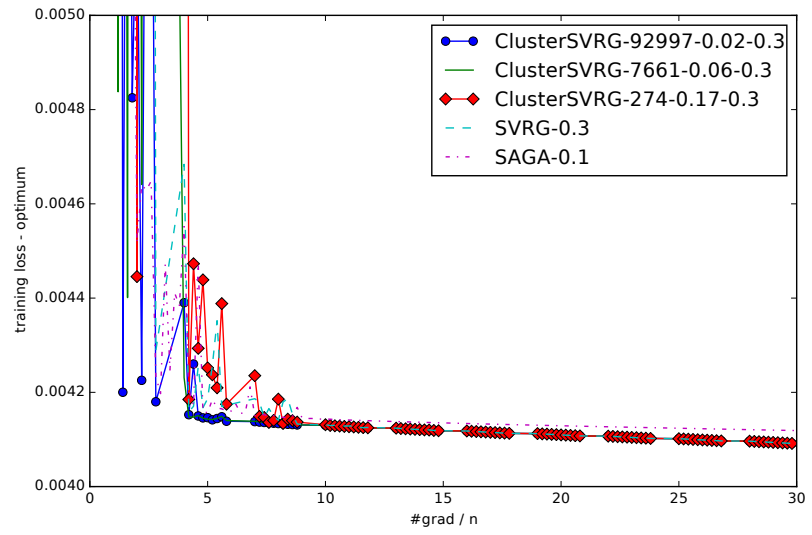

(f) Covtype, Lasso $\lambda = 10^{-5}$

Figure 2: Comparing ClusterSVRG with SVRG, SAGA on Covtype

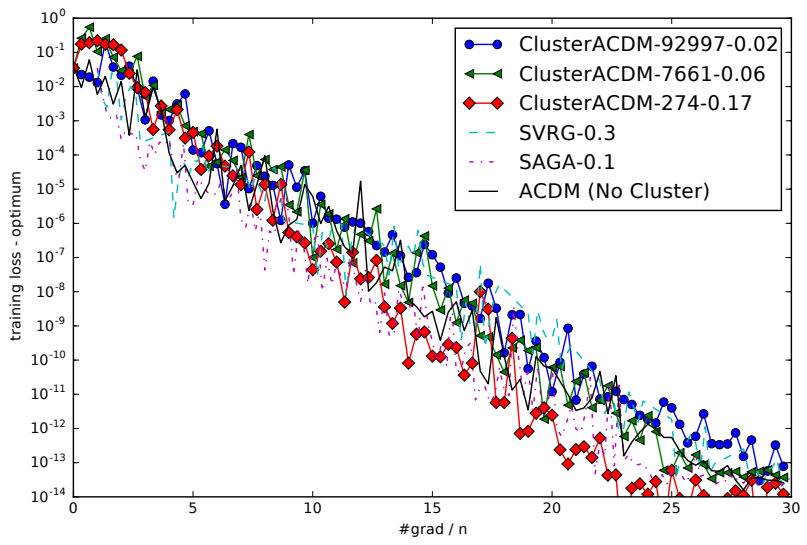

(a) Covtype, Ridge $\lambda = 10^{-5}$

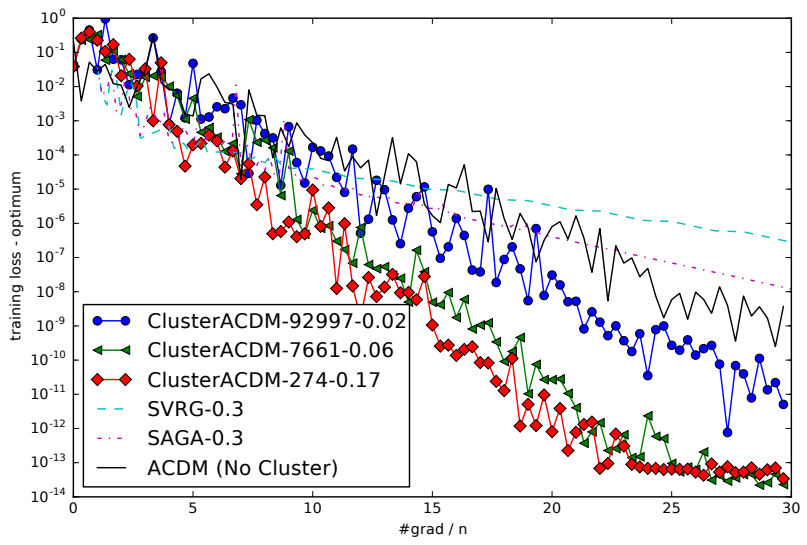

(b) Covtype, Ridge $\lambda = 10^{-6}$

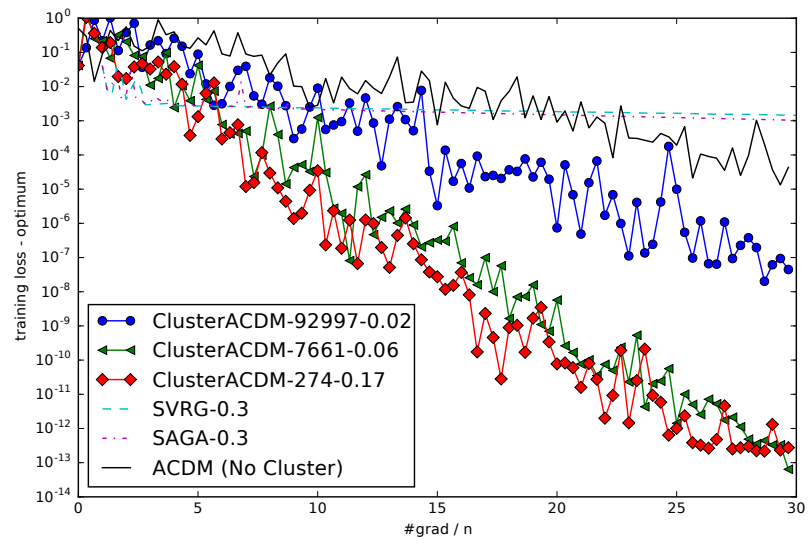

(c) Covtype, Ridge $\lambda = 10^{-7}$

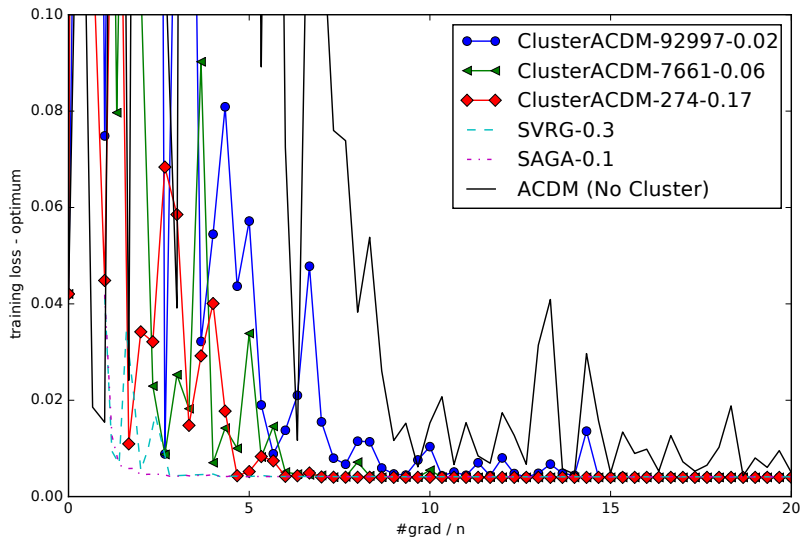

(d) Covtype, Lasso $\lambda = 10^{-5}$

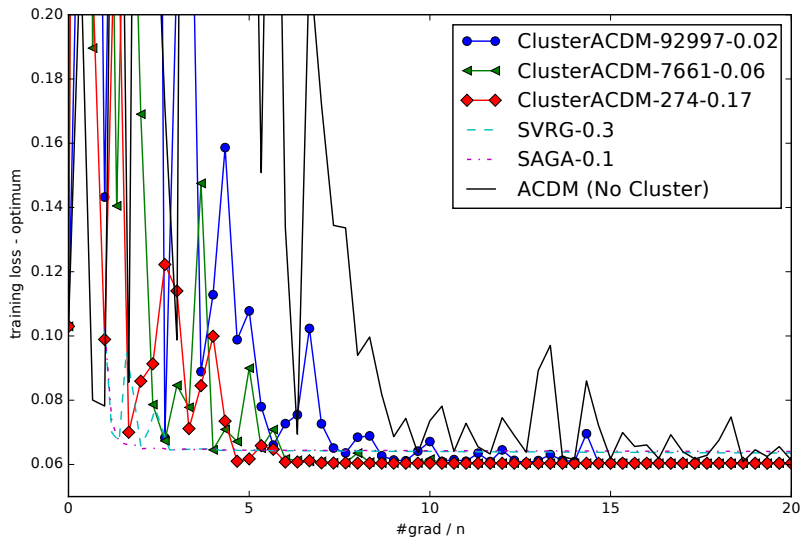

(e) Covtype, Lasso $\lambda = 10^{-6}$

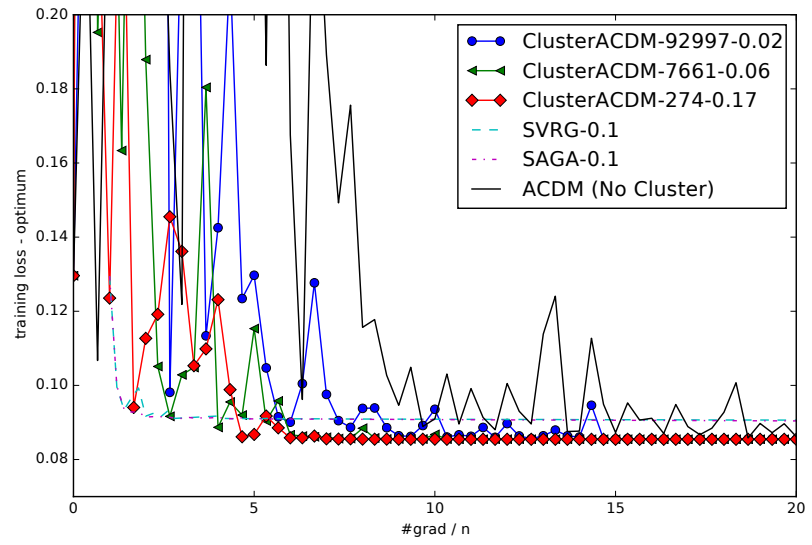

(f) Covtype, Lasso $\lambda = 10^{-7}$

Figure 3: Comparing ClusterACDM with SAGA, SVRG, ACDM on Covtype

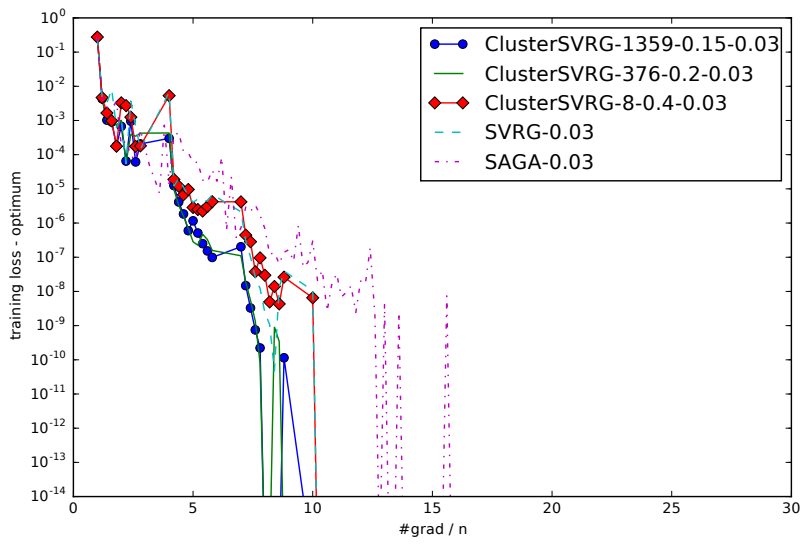

(a) SensIT, Ridge $\lambda = 10^{-3}$

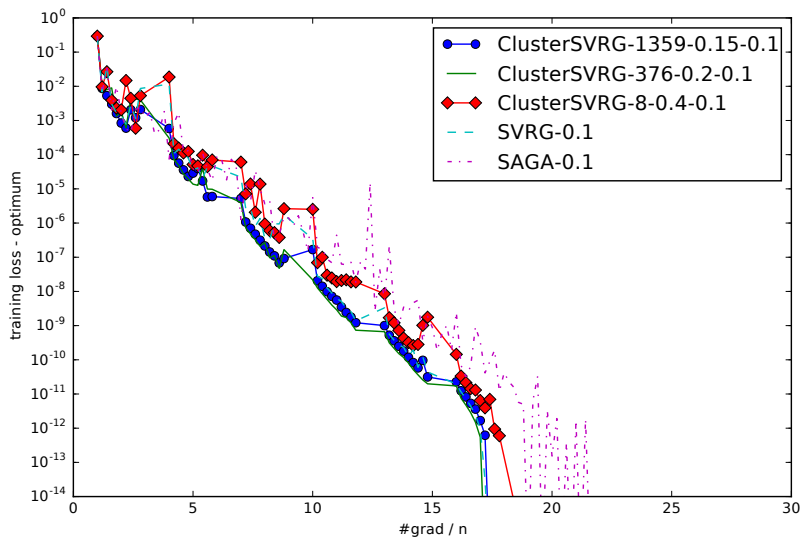

(b) SensIT, Ridge $\lambda = 10^{-4}$

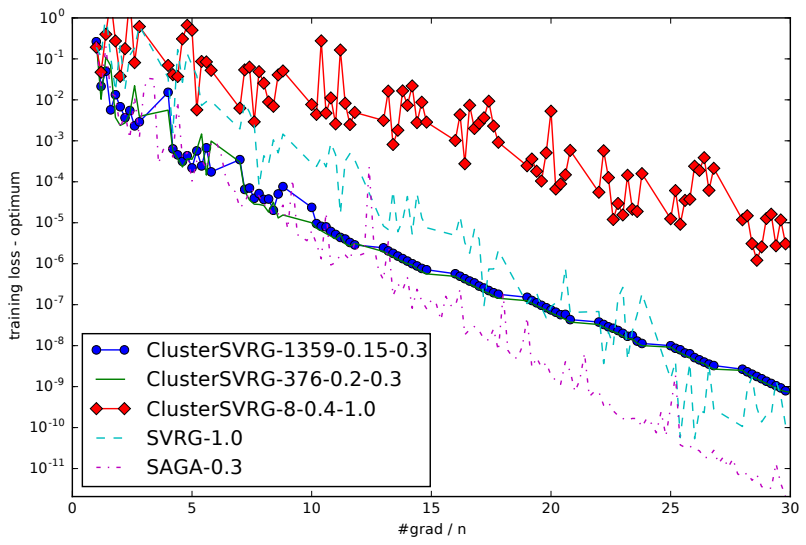

(c) SensIT, Ridge $\lambda = 10^{-5}$

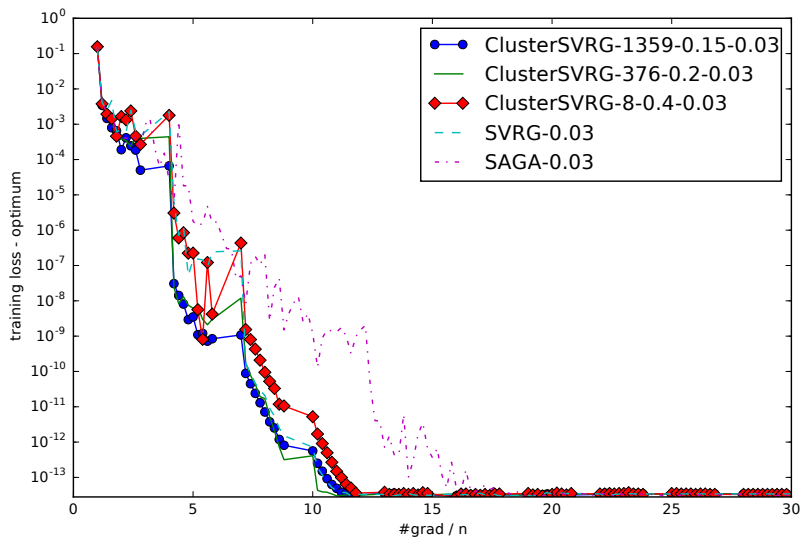

(d) SensIT, Lasso $\lambda = 10^{-2}$

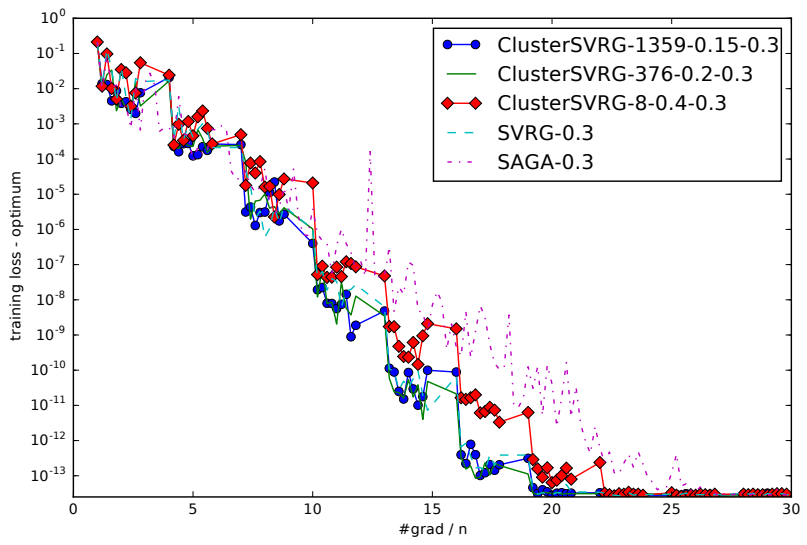

(e) SensIT, Lasso $\lambda = 10^{-3}$

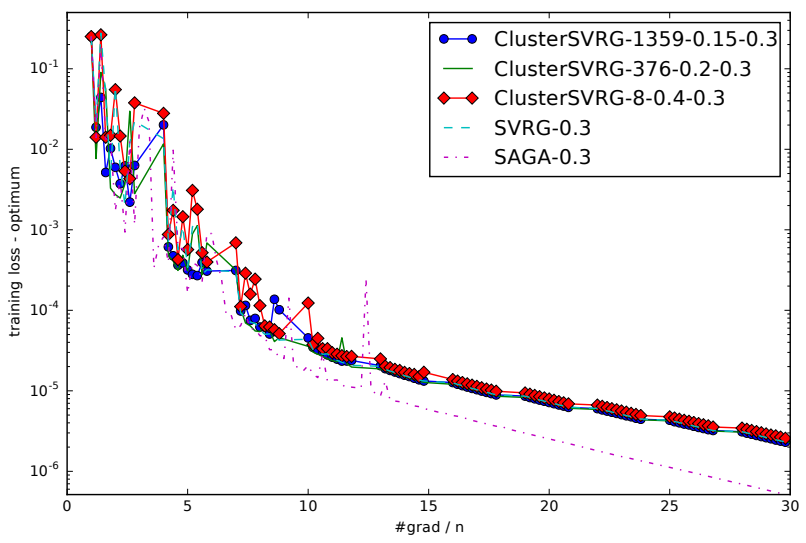

(f) SensIT, Lasso $\lambda = 10^{-4}$

Figure 4: Comparing ClusterSVRG with SVRG, SAGA on SensIT

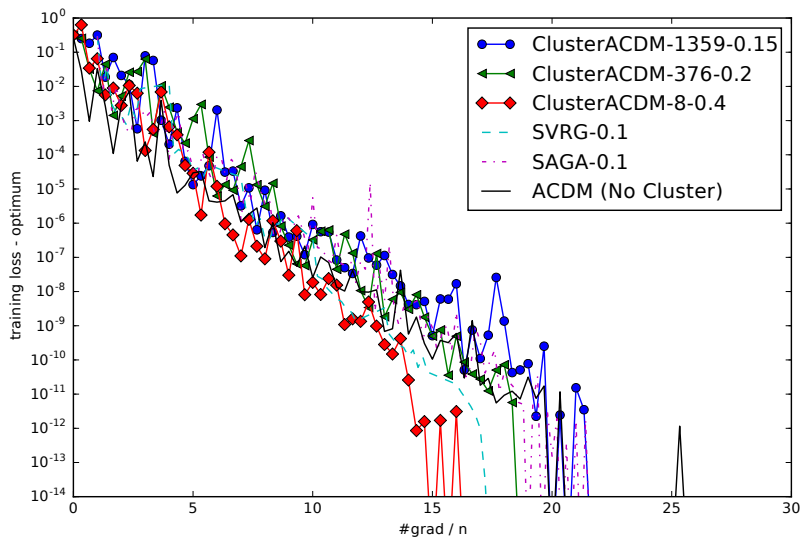

(a) SensIT, Ridge $\lambda = 10^{-4}$

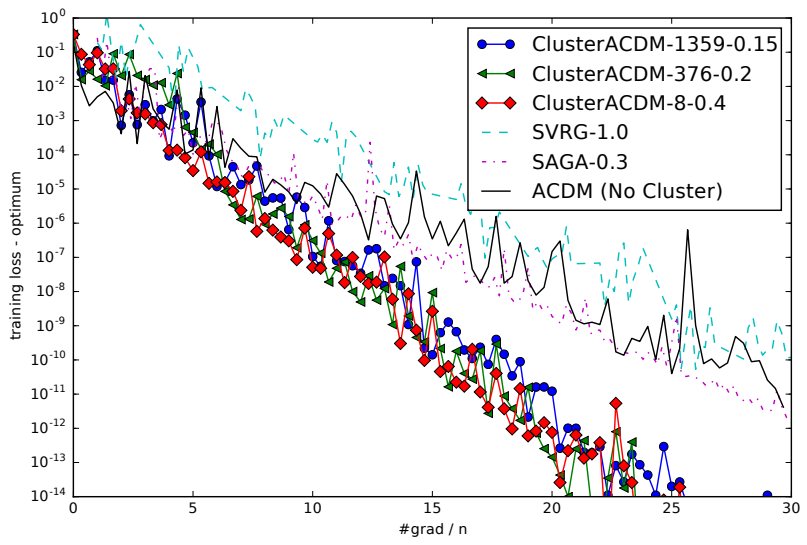

(b) SensIT, Ridge $\lambda = 10^{-5}$

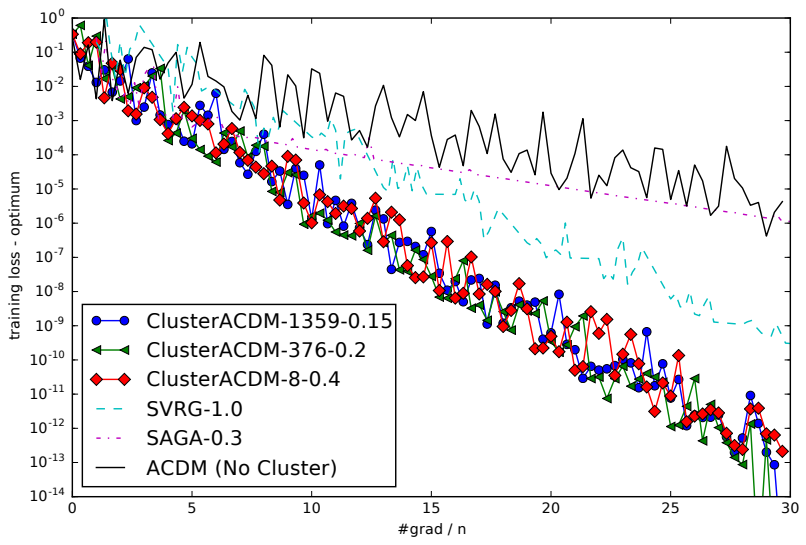

(c) SensIT, Ridge $\lambda = 10^{-6}$

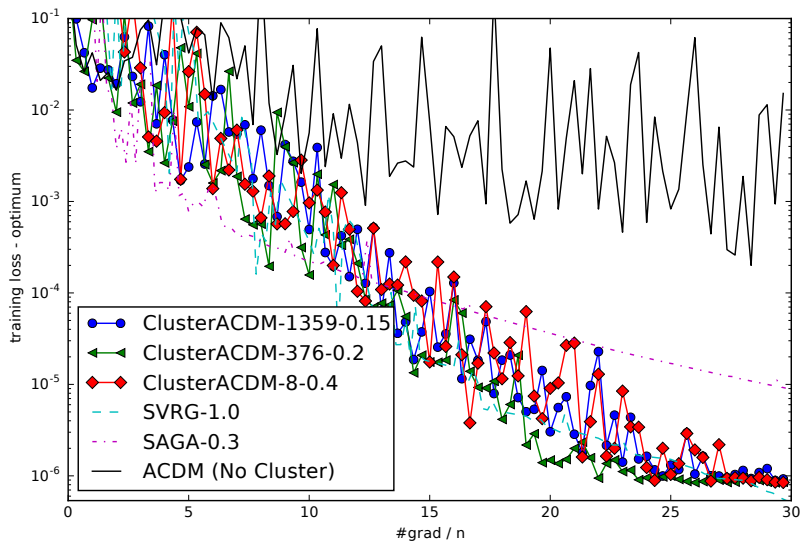

(d) SensIT, Lasso $\lambda = 10^{-5}$

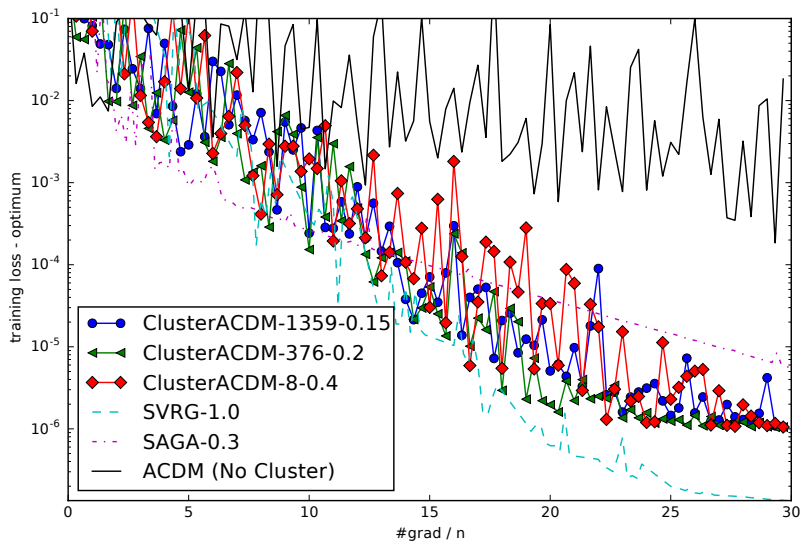

(e) SensIT, Lasso $\lambda = 10^{-6}$

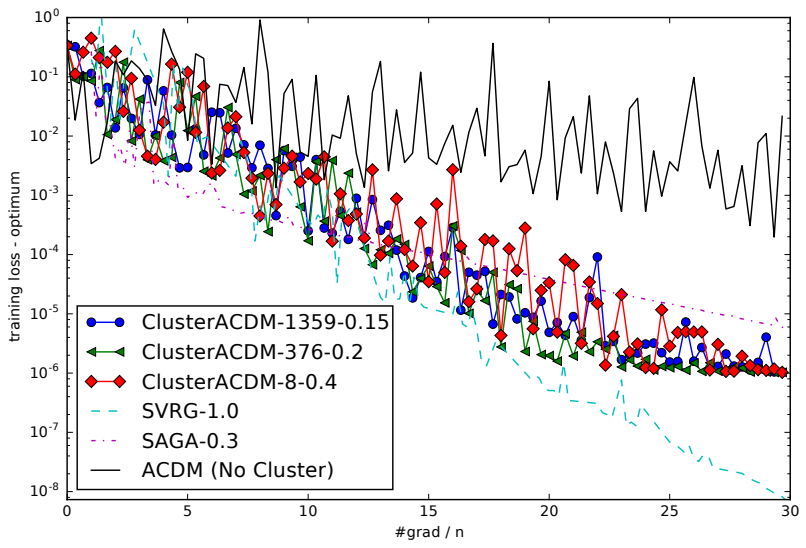

(f) SensIT, Lasso $\lambda = 10^{-7}$

Figure 5: Comparing ClusterACDM with SVRG, SAGA, ACDM on SensIT

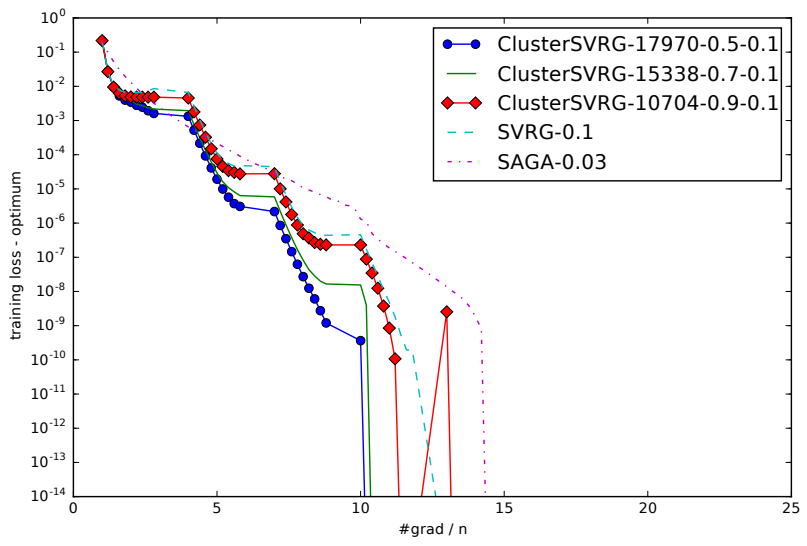

(a) News20, Ridge $\lambda = 10^{-3}$

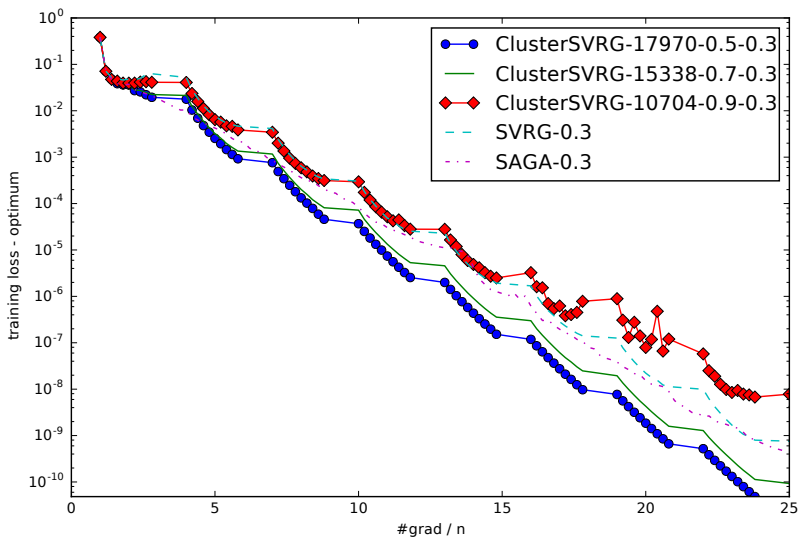

(b) News20, Ridge $\lambda = 10^{-4}$

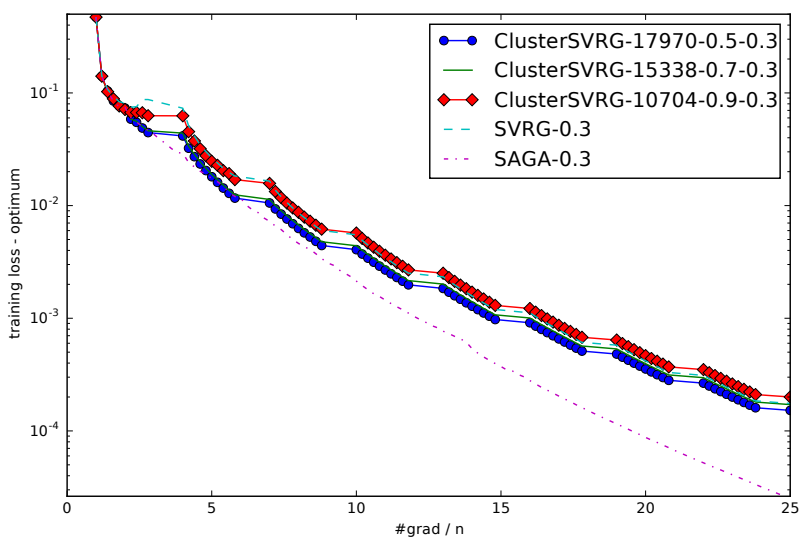

(c) News20, Ridge $\lambda = 10^{-5}$

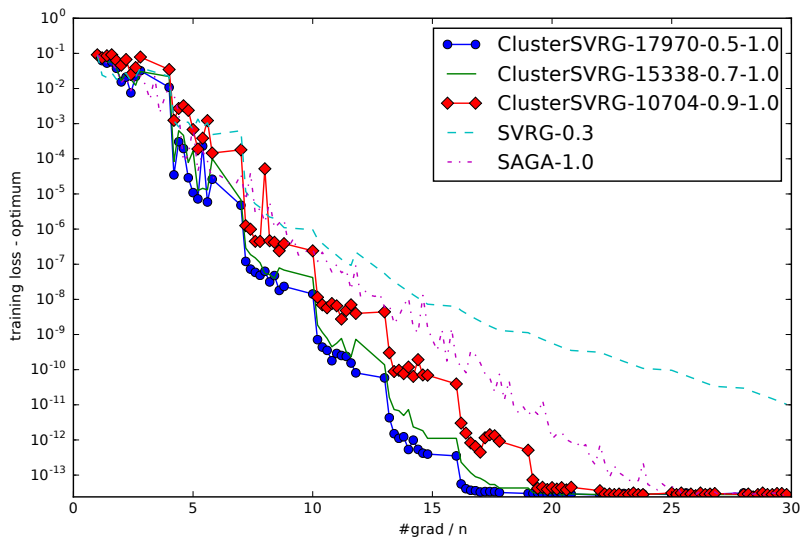

(d) News20, Lasso $\lambda = 10^{-3}$

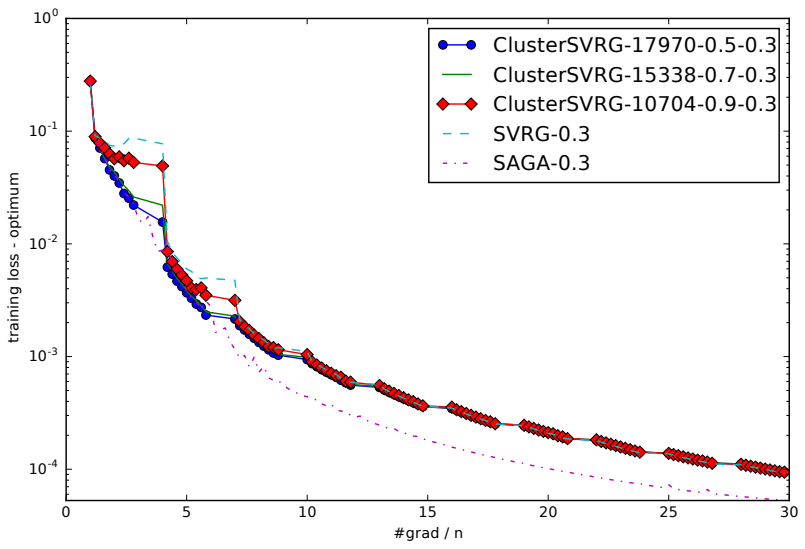

(e) News20, Lasso $\lambda = 10^{-4}$

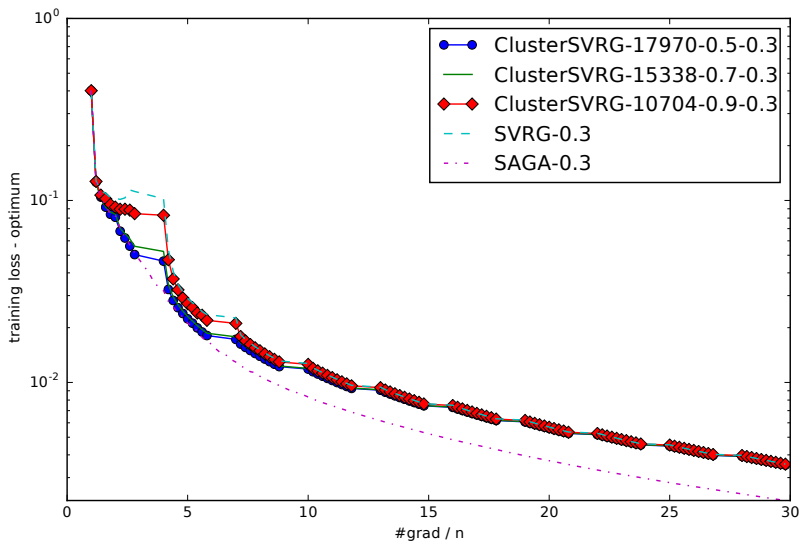

(f) News20, Lasso $\lambda = 10^{-5}$

Figure 6: Comparing ClusterSVRG with SVRG, SAGA on News20

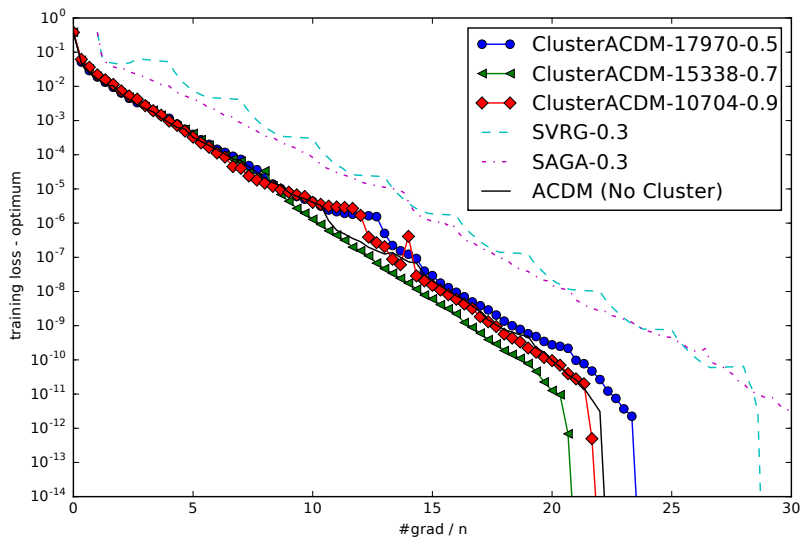

(a) News20, Ridge $\lambda = 10^{-4}$

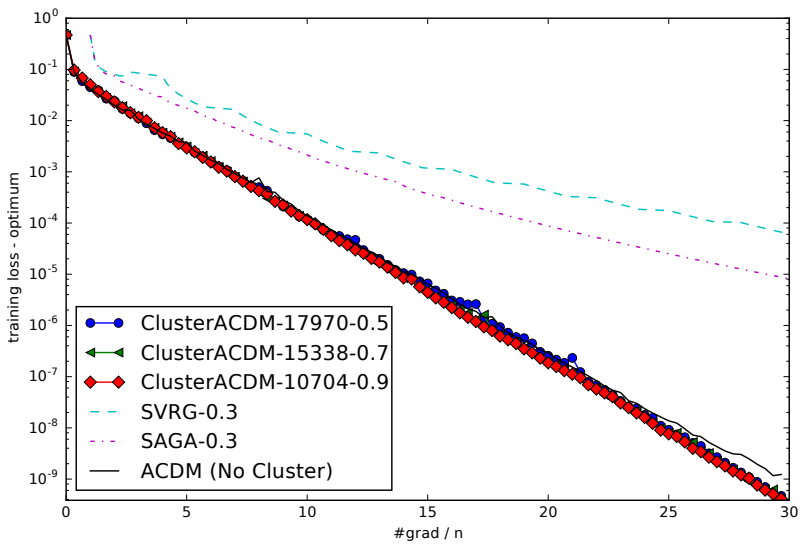

(b) News20, Ridge $\lambda = 10^{-5}$

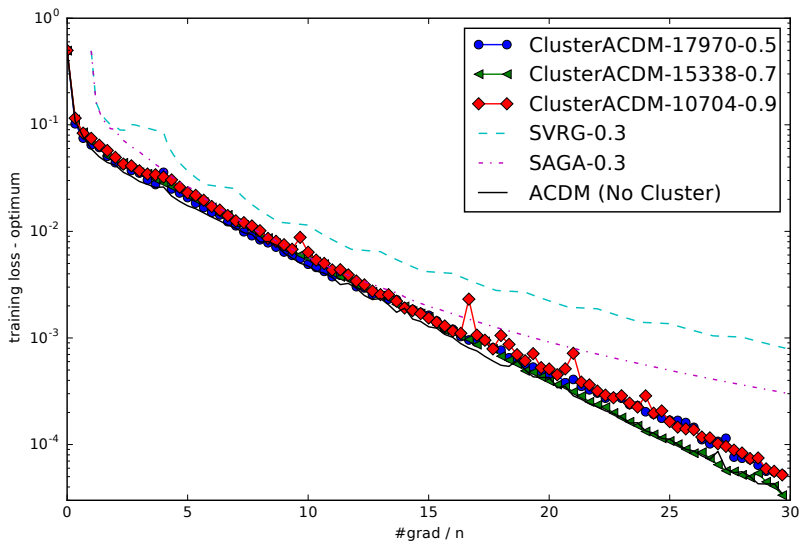

(c) News20, Ridge $\lambda = 10^{-6}$

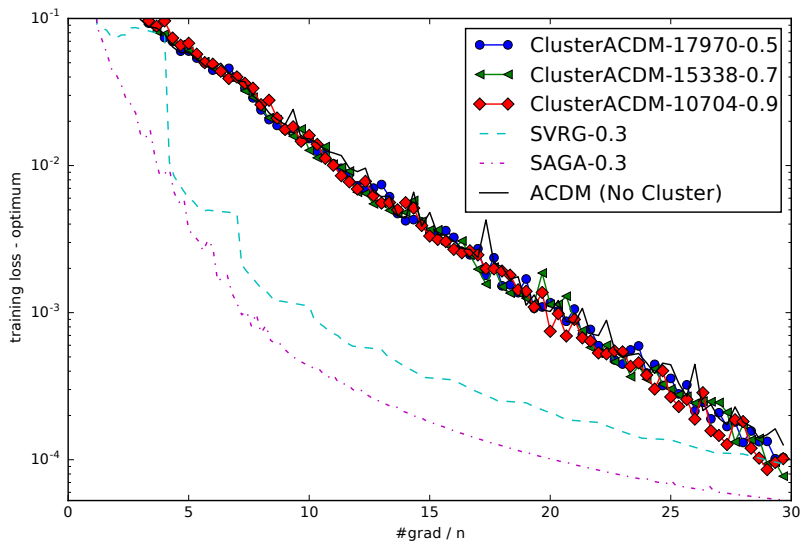

(d) News20, Lasso $\lambda = 10^{-4}$

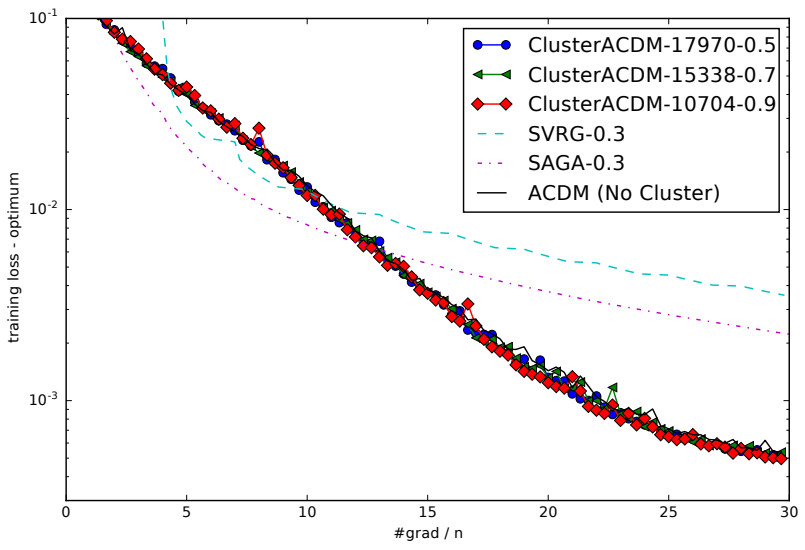

(e) News20, Lasso $\lambda = 10^{-5}$

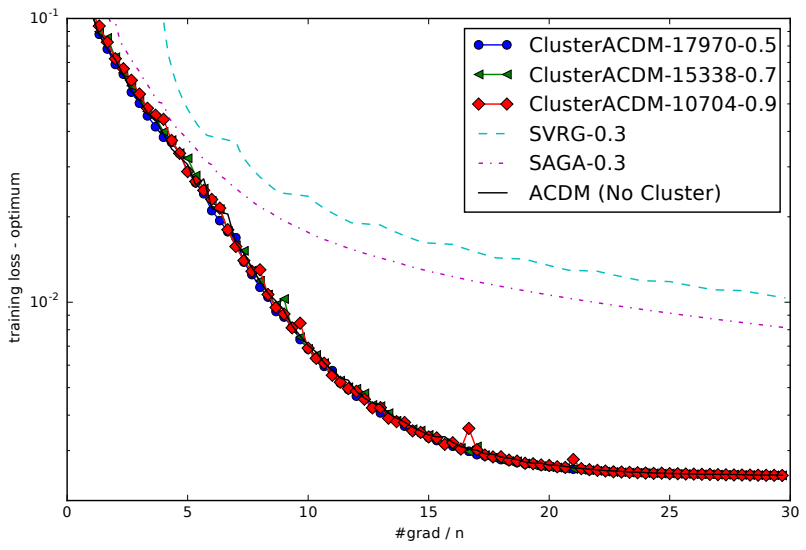

(f) News20, Lasso $\lambda = 10^{-6}$

Figure 7: Comparing ClusterACDM with SVRG, SAGA, ACDM on News20

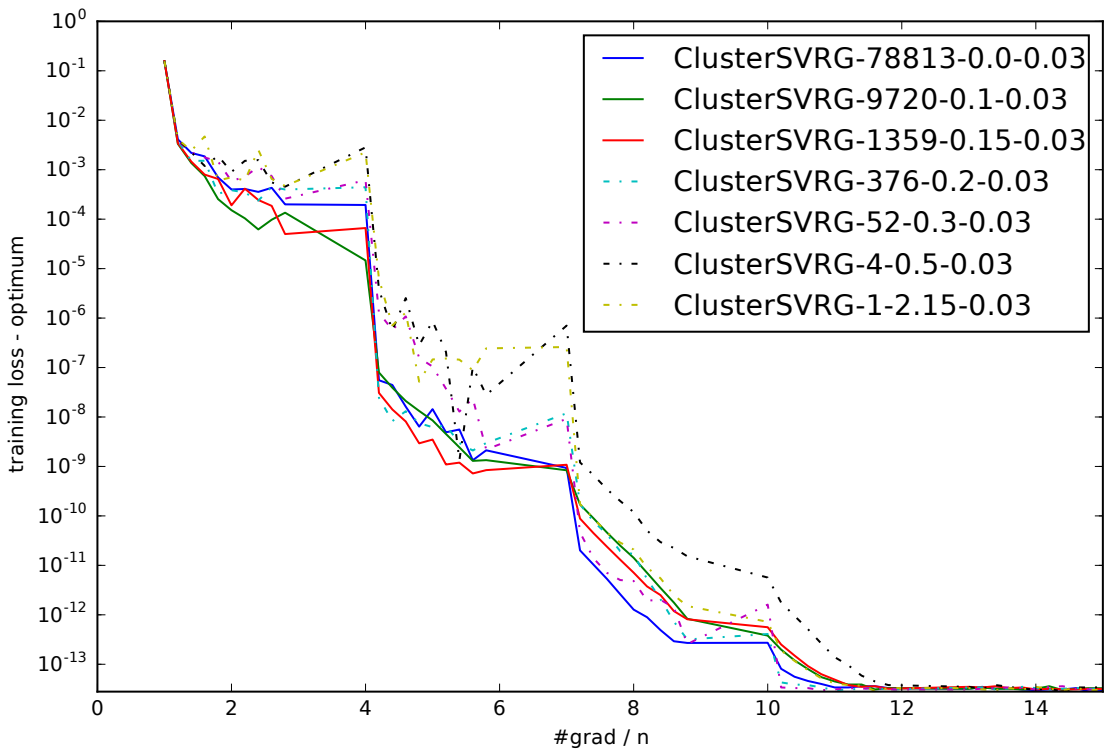

(a) SensIT, Lasso $\lambda = 10^{-2}$

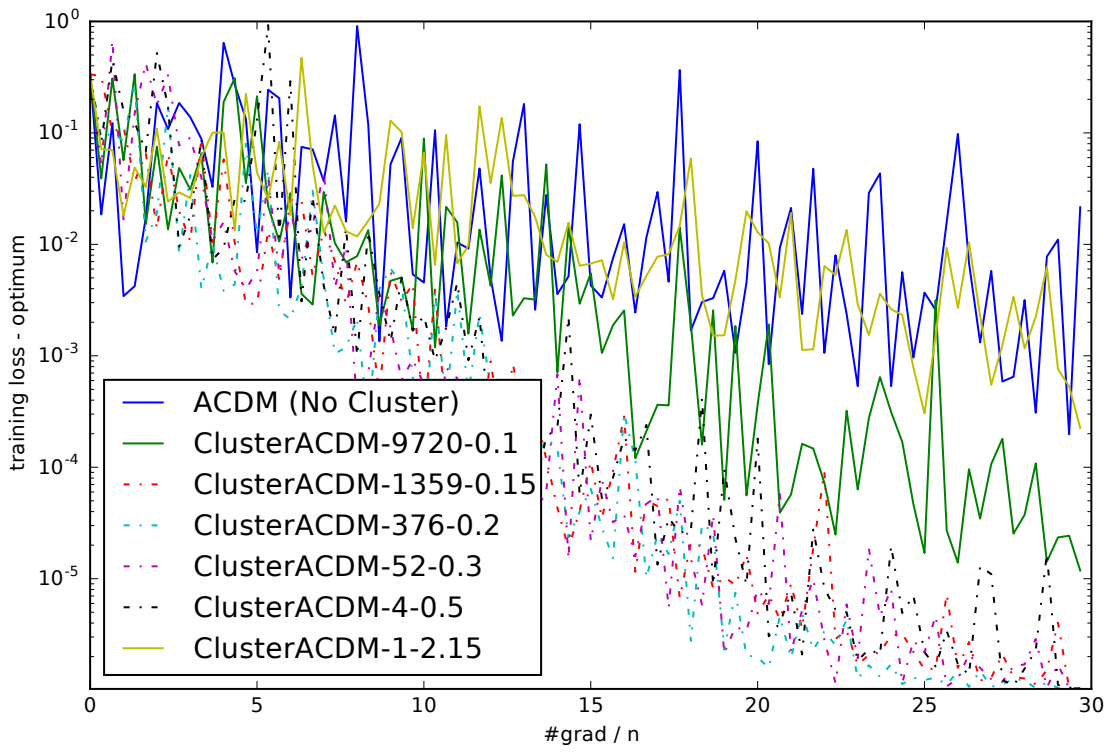

(b) SensIT, Lasso $\lambda = 10^{-7}$

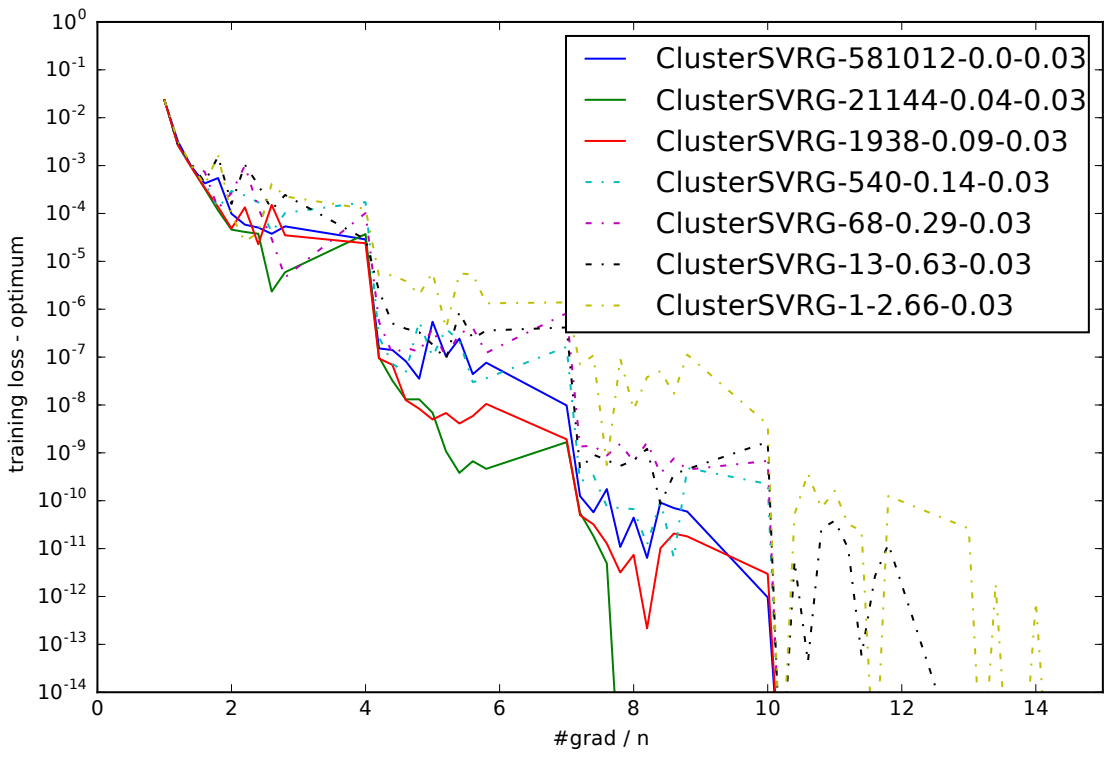

(c) Covtype, Ridge $\lambda = 10^{-4}$

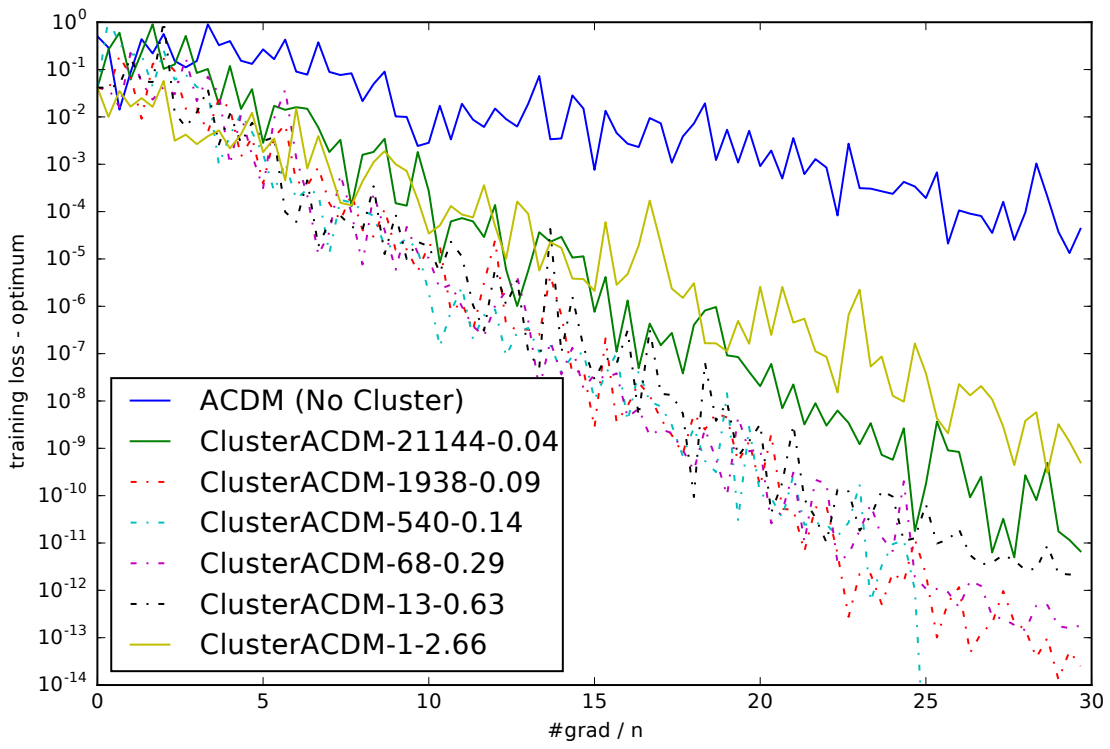

(d) Covtype, Ridge $\lambda = 10^{-7}$

Figure 8: Full clustering

# A Convergence Analysis for ClusterACDM

## A.1 Missing Remark, Lemma, Examples

We first state some missing remark, lemma, and example that we excluded from our main body of this paper due to space limitation.

**Remark A.1.** If $r(x)$ is not strongly convex, we can always make $r(x)$ it strongly convex by a standard reduction. For instance, in Lasso we may have $r(x) = \lambda_1 \|x\|_1$ in the primal objective (3.1) which is not strongly convex. In this case, it suffices to look at an alternative regularizer $r'(x) \stackrel{\text{def}}{=} r(x) + \frac{\lambda}{2}\|x\|_2^2$ and its corresponding objective function $P'(x) \stackrel{\text{def}}{=} P(x) + \frac{\lambda}{2}\|x\|_2^2$. Since for every $x$ it satisfies $|P(x) - P'(x)| \leq O(\lambda)$, one can specify a small enough parameter $\lambda$ proportional to the desired error $\varepsilon$, and minimize $P'(x)$ instead. This dummy 2-norm regularizer introduces error to the objective, but allows the function $P(x)$ to be minimized in $\widetilde{O}(nd + \sqrt{n/\lambda}d) = \widetilde{O}(nd + \sqrt{n/\varepsilon d})$ time, again matching the fastest known running time by AccSDCA. (In fact, all accelerated stochastic methods need to introduce dummy $\ell_2$ regularizers.) Thus, in analyzing ClusterACDM, it suffices for us to analyze the case when $r(x)$ is $\lambda$-strongly convex for some $\lambda > 0$.

**Lemma 3.1** (restated)**.** *If $r(x)$ is $\lambda$-strongly convex, then $D(y)$ is $\sigma = \frac{1}{n}$ strongly convex and coordinate-wise smooth with parameters $(L_1, \ldots, L_n)$ for $L_i = \frac{1}{n} + \frac{1}{\lambda n^2}\|a_i\|^2$.*

*Proof.* Since there is a component $\frac{1}{2n}\|y\|^2$ in the definition of $D(y)$, we know that $D(y)$ is $\frac{1}{n}$-strongly convex. To prove the smoothness, recall that if $r(\cdot)$ is $\lambda$ strongly convex, then $r^*(\cdot)$ is $1/\lambda$ smooth, or mathematically, $\nabla^2 r^*(x) \preceq \frac{1}{\lambda} I$ (see for instance the text book [5]). Here we denote by $A \preceq B$ the spectral dominance of matrices, which means $x^T A x \leq x^T B x$ for every $x \in \mathbb{R}^n$.

Using this fact, we compute the Hessian of $D(\cdot)$

$$\nabla^2 D(y) = \frac{1}{n} I + \nabla^2 r^*\big(-\frac{1}{n} A y\big) \preceq \frac{1}{n} I + \frac{1}{n^2} A^T \frac{I}{\lambda} A = \frac{1}{n} + \frac{1}{\lambda n^2} A^T A \ .$$

Now, since $(A^T A)_{ii} = \|a_i\|^2$ for each $i \in [n]$, we have that $\nabla^2_{ii} D(y) \leq \frac{1}{n} + \frac{1}{\lambda n^2}\|a_i\|^2$ which gives the desired smoothness property. □

We give a few examples of the Haar matrices we introduced in Definition 3.4.

**Example A.2.** We have

$$H_4 = \begin{bmatrix} 1/2 & 1/2 & 1/2 & 1/2 \\ 1/2 & 1/2 & -1/2 & -1/2 \\ 1/\sqrt{2} & -1/\sqrt{2} & 0 & 0 \\ 0 & 0 & 1/\sqrt{2} & -1/\sqrt{2} \end{bmatrix}$$

and

$$H_7 = \begin{bmatrix} 1/\sqrt{7} & 1/\sqrt{7} & 1/\sqrt{7} & 1/\sqrt{7} & 1/\sqrt{7} & 1/\sqrt{7} & 1/\sqrt{7} \\ 2/\sqrt{21} & 2/\sqrt{21} & 2/\sqrt{21} & -\sqrt{21}/14 & -\sqrt{21}/14 & -\sqrt{21}/14 & -\sqrt{21}/14 \\ \sqrt{2}/\sqrt{3} & -\sqrt{2}/(2\sqrt{3}) & -\sqrt{2}/(2\sqrt{3}) & 0 & 0 & 0 & 0 \\ 0 & 1/\sqrt{2} & -1/\sqrt{2} & 0 & 0 & 0 & 0 \\ 0 & 0 & 0 & 1/2 & 1/2 & -1/2 & -1/2 \\ 0 & 0 & 0 & 1/\sqrt{2} & -1/\sqrt{2} & 0 & 0 \\ 0 & 0 & 0 & 0 & 0 & 1/\sqrt{2} & -1/\sqrt{2} \end{bmatrix}$$

## A.2 Proof Details

The following statement is a simple corollary of Lemma 3.1 on the transformed objective:

**Corollary A.3.** *Letting $b_1, \dots, b_n \in \mathbb{R}^d$ be the column vectors of $AH^T$, then the transformed objective $D'(\cdot)$ is coordinate-wise smooth with parameters $(L'_1, \dots, L'_n)$ for $L'_i = \frac{1}{n} + \frac{1}{\lambda n^2}\|b_i\|^2$.*

We are now ready to relate the smoothness parameters of the transformed objective to how good a raw clustering is. Suppose that $\|a_1\|^2 = \cdots \|a_n\|^2 = 1$ for the sake of simplicity.[13] We consider the following definition of raw clustering:

**Definition A.4** (raw clustering on vectors)**.** *We say that a partition $S = S_1 \cup \cdots \cup S_s$ is a $(s, \delta)$ raw clustering, if for every cluster $c$, it satisfies that $\frac{1}{\|S_c\|^2}\sum_{i,j\in S_c}\langle a_i, a_j\rangle \ge 1 - \frac{\delta}{2}$. In other words, in average, the pairwise inner products of features vectors in each cluster $S_c$ are at least $1 - \frac{\delta}{2}$.*

Since $\|a_i - a_j\|^2 = 2 - 2\langle a_i, a_j\rangle$, the above definition is equivalent to Definition 2.1. It implies the following property on the coordinate smoothness of the transformed objective:

**Lemma A.5.** *If the transformed objective is built on a $(s, \delta)$ raw clustering, then it satisfies*

$$\sum_{i\in[n]}\sqrt{L'_i} \le \sqrt{n} + \frac{\sqrt{c/n + \delta/2}}{\sqrt{\lambda}}$$

*Proof.* For each cluster $c \in [s]$, recall that the first row of the Haar matrix $H_{n_c}$ is all $1/\sqrt{n_c}$. Let $i_c \in [n]$ be the index of this first row in our matrix $H$ for each cluster $c$, which are "*important*" indices. Since $b_{i_c}$ is the $i$-th column vector of $AH^T = [a_1, \dots, a_n]H^T$, we have

$$b_{i_c} = \frac{1}{\sqrt{n_c}}\sum_{j\in S_c} a_j \ .$$

Therefore, we have $\|b_{i_c}\|^2 = \frac{1}{n_c}\sum_{j_1,j_2\in S_c}\langle a_{j_1}, a_{j_2}\rangle \ge (1 - \delta/2)n_c$ due to our definition of the $(s, \delta)$ raw clustering. Summing them up, we have

$$\sum_{i\in\{i_1,\dots,i_s\}}\|b_i\|^2 \ge \Big(1 - \delta/2\Big)n$$

and we define $\delta' \le \delta$ to be the parameter that satisfies $\sum_{i\in\{i_1,\dots,i_s\}}\|b_i\|^2 = (1 - \frac{\delta'}{2})n$.

Now we compute the summation of $\sqrt{L'_i}$ by dividing indices into important indices $i_1, i_2, \dots, i_s$ and others. We first sum them up for important indices:

$$\sum_{i\in\{i_1,\dots,i_s\}}\sqrt{L'_i} \le \sqrt{s}\cdot\Big(\sum_{i\in\{i_1,\dots,i_s\}}L'_i\Big)^{1/2} = \sqrt{s}\cdot\Big(\frac{s}{n} + \frac{(1-\frac{\delta'}{2})n}{\lambda n^2}\Big)^{1/2}$$

On the other hand, for non-important indices, we have

$$\sum_{i\notin\{i_1,\dots,i_s\}}\sqrt{L'_i} \le \sqrt{n-s}\cdot\Big(\sum_{i\notin\{i_1,\dots,i_s\}}L'_i\Big)^{1/2} = \sqrt{n-s}\cdot\Big(\frac{n-s}{n} + \frac{\delta' n}{2\lambda n^2}\Big)^{1/2}$$

Together, we have

$$\sum_{i\in[n]}\sqrt{L'_i} \le \sqrt{s}\cdot\sqrt{\frac{s}{n}} + \sqrt{n-s}\cdot\sqrt{\frac{n-s}{n}} + \sqrt{s}\cdot\sqrt{\frac{1-\frac{\delta'}{2}}{\lambda n}} + \sqrt{n-s}\cdot\sqrt{\frac{\delta'}{2\lambda n}} \le \sqrt{n} + \frac{\sqrt{s/n+\delta/2}}{\sqrt{\lambda}} \ . \quad \square$$

[13] We remark here that in some machine learning applications, the feature vectors are already normalized so this already holds. If not, one can treat vectors with different magnitudes of norms separately and perform raw clustering.

Finally, using the fact that $D'(\cdot)$ is $\sigma = \frac{1}{n}$ strongly convex, we can apply Theorem 2.3 and deduce Theorem 3.5.

# B ClusterSVRG Algorithm

In this section we more formally introduce our ClusterSVRG method.

While our accelerated method ClusterACDM works only for regularized least-square problems in theory, our non-accelerated method ClusterSVRG, like its parent method SVRG [13], focuses on a more general stochastic setting, consisting of a set of $n$ convex functions $\{f_1(x), \dots, f_n(x)\}$ rather than $n$ vectors. In this case, we consider the following quantification of the clustering quality:

**Definition B.1** (raw clustering on functions)**.** *We say a partition $[n] = S_1 \cup \cdots \cup S_s$ is an $(s, \sigma)$ raw clustering of the functions $\{f_1(x), \dots, f_n(x)\}$ if for every cluster $S_c$ and every $x$ that has bounded norms, it satisfies*

$$\frac{1}{|S_c|^2} \sum_{i,j \in S_c} \|\nabla f_i(x) - \nabla f_j(x)\|^2 \leq \sigma \enspace .$$

**Remark B.2.** Definition B.1 is a natural generalization of Definition 2.1 for the following reason. Consider the case when each function $f_i(x) = \frac{1}{2}(\langle a_i, x\rangle - 1)^2$ is a least square loss function, where $a_i$ is the a feature vector with $\|a_i\| \leq 1$. Then, it satisfies that $\nabla f_i(x) = (\langle a_i, x\rangle - 1)a_i$ and therefore

$$\begin{aligned}
&\|\nabla f_i(x) - \nabla f_j(x)\|^2 = \|\langle a_i, x\rangle a_i - \langle a_j, x\rangle a_j - a_i + a_j\|^2 \\
&= \|\big(\langle a_i, x\rangle - 1\big)(a_i - a_j) + \langle a_i - a_j, x\rangle a_j\|^2 \\
&\leq 2\Big(\big(\langle a_i, x\rangle - 1\big)^2 \cdot \|a_i - a_j\|^2 + \big(\langle a_i - a_j, x\rangle\big)^2 \cdot \|a_j\|^2\Big) \\
&\leq O\big(\|x\|^2\big) \cdot \|a_i - a_j\|^2 \enspace .
\end{aligned}$$

Since we have assumed $x$ to be bounded, it follows that an $(s, \delta)$ raw clustering on the dataset implies an $(s, \sigma)$ raw clustering on the corresponding least-square loss functions where the parameter $\sigma$ is on the same order as $\delta$. Similar results can be deduced for other loss functions including logistic loss, smoothed hinge losses, etc.

## B.1 Previous Solutions

Variance-reduction based stochastic gradient methods iteratively perform the following update

$$x^t = \arg\min_x \Big\{ \frac{1}{2\eta}\|x - x^{t-1}\|^2 + \langle \widetilde{\nabla}^{t-1}, x\rangle + \Psi(x) \Big\} \enspace ,$$

where $\widetilde{\nabla}^{t-1}$ is the so-called gradient estimator and usually its expectation has to equal to the full gradient $\nabla f(x^{t-1})$ in order to make the algorithm *unbiased.* The central idea behind all such methods is to ensure that the variance $\mathbf{Var}[\widetilde{\nabla}^{t-1}]$ decreases quickly.

In particular, SVRG has an outer loop of epochs, where at the beginning of each epoch, SVRG records the position of the current iterate $x$ as the snapshot point $\widetilde{x}$, [14]and computes its full gradient $\nabla f(\widetilde{x})$. Each epoch consists of $m$ inner iterations, and in each inner iteration, SVRG picks a random $i$ and defines the gradient estimator

$$\widetilde{\nabla}^{t-1}_{\mathrm{SVRG}} \stackrel{\mathrm{def}}{=} \frac{1}{n}\sum_{j=1}^{n} \nabla f_j(\widetilde{x}) + \nabla f_i(x^{t-1}) - \nabla f_i(\widetilde{x}) \enspace . \tag{B.1}$$

[14]More precisely, SVRG provides two options, one defining $\widetilde{x}$ as the average iterates of the previous epoch, and the other defining $\widetilde{x}$ to be the last iterate of the previous epoch.

SAGA works differently from SVRG. It maintains a table of $n$ vectors that stores the gradient $\nabla f_i(\phi_i)$ at position $\phi_i$ for each $i \in [n]$. In each iteration $t$, SAGA picks a random $i$ and define

$$\widetilde{\nabla}^{t-1}_{\mathrm{SAGA}} \stackrel{\mathrm{def}}{=} \frac{1}{n}\sum_{j=1}^{n} \nabla f_j(\phi_j) + \nabla f_i(x^{t-1}) - \nabla f_i(\phi_i) \enspace .$$

After updating $x^t$ using this estimator, SAGA records this new position $\phi_i \leftarrow x^{t-1}$ and update the corresponding $\nabla f_i(\phi_i)$ in the table.

## B.2 Our New Algorithm

Our ClusterSVRG computes the gradient estimator $\widetilde{\nabla}$ based on the clustering information. Given a clustering $[n] = S_1 \cup \cdots \cup S_s$, we denote by $c_i \in [s]$ the cluster that index $i$ belongs to. Using the same snapshot notation $\widetilde{x}$ as SVRG, we define

$$\widetilde{\nabla}^{t-1} \stackrel{\mathrm{def}}{=} \frac{1}{n}\sum_{j=1}^{n} \big(\nabla f_j(\widetilde{x}) + \zeta_{c_j}\big) + \nabla f_i(x^{t-1}) - \big(\nabla f_i(\widetilde{x}) + \zeta_{c_i}\big) \tag{B.2}$$

Above, $i$ is chosen uniformly at random from $[n]$, and for each cluster $c$, we introduce an additional $\zeta_c$ that we shall define shortly. The existence of $\zeta_c$ for different clusters makes our $\widetilde{\nabla}^{t-1}$ different from $\widetilde{\nabla}^{t-1}_{\mathrm{SVRG}}$. However, if there were only one cluster, then $\widetilde{\nabla}^{t-1}$ would be identical to $\widetilde{\nabla}^{t-1}_{\mathrm{SVRG}}$ no matter how $\zeta_c$ is defined.

We note that regardless of the definition of $\zeta_c$, it satisfies $\mathbb{E}_i[\zeta_{c_i}] = \sum_{c=1}^{s} \frac{n_c}{n}\zeta_c = \sum_{j=1}^{n} \frac{\zeta_{c_j}}{n}$ and therefore our estimator is unbiased: that is, $\mathbb{E}_i[\widetilde{\nabla}^{t-1}] = \nabla f(x^{t-1})$.

Similar to SVRG, we propose two slightly different definitions for $\zeta_c$, and call them Option I and Option II. In both options, we initialize $\zeta_c = 0$ at the beginning of the epoch.

- In Option I, after each iteration $t$ is completed and suppose $i$ is the random index chosen at iteration $t$, we update $\zeta_{c_i} \leftarrow \nabla f_i(x^{t-1}) - \nabla f_i(\widetilde{x})$.
- In Option II, we divide an epoch of length $m$ into subepochs of length $s$ each (recall $s$ is the number of clusters). At the beginning of each subepoch, for each cluster $c \in [s]$, we define $\zeta_c \leftarrow \nabla f_j(\overline{x}) - \nabla f_j(\widetilde{x})$. Here, $\overline{x}$ is the last iterate of the previous subepoch and $j$ is a random index in $S_c$.

We summarize both options in Algorithm 2.

**High Level Intuition.** In the estimator $\widetilde{\nabla}^{t-1}_{\mathrm{SVRG}}$, each $\nabla f_j(\widetilde{x})$ intuitively serves as a "rough guess" of the gradient $\nabla f_j(x^{t-1})$ with respect to the iterate $x^{t-1}$. However, this guess may be quite different from $\nabla f_j(x^{t-1})$ because at most $m$ iterations may have passed between $\widetilde{x}$ and $x^{t-1}$. In this case, we say $\nabla f_j(\widetilde{x})$ has an *outdatedness* of $O(m)$. Since the epoch length $m$ is usually of the same magnitude as $n$ [13], this outdatedness may be very large. Our newly introduced $\zeta_c$, as a cluster-dependent correction term, will help us decrease this outdatedness.

More specifically, in Option I, define $\mathsf{lt}(c)$ to be the most recent time $t$ when any index $i$ in cluster $S_c$ was randomly chosen, and define $\mathsf{li}(c)$ to be this index $i$. Then, we can write

$$\zeta_c = \nabla f_{\mathsf{li}(c)}(x^{\mathsf{lt}(c)}) - \nabla f_{\mathsf{li}(c)}(\widetilde{x}) \enspace .$$

Under this notation, we claim that $\nabla f_j(\widetilde{x}) + \zeta_{c_j}$ is a better guess of $\nabla f_j(x^{t-1})$ comparing to $\nabla f_j(\widetilde{x})$. Indeed, if all the gradients $\nabla f_k(x)$ for $k \in S_{c_j}$ are close enough (which follows from the raw clustering definition in Definition B.1), then $\nabla f_j(\widetilde{x}) \approx \nabla f_{\mathsf{li}(c_j)}(\widetilde{x})$ and therefore

$$\nabla f_j(\widetilde{x}) + \zeta_{c_j} = \nabla f_j(\widetilde{x}) + \nabla f_{\mathsf{li}(c_j)}(x^{\mathsf{lt}(c_j)}) - \nabla f_{\mathsf{li}(c_j)}(\widetilde{x}) \approx \nabla f_{\mathsf{li}(c_j)}(x^{\mathsf{lt}(c_j)}) \approx \nabla f_j(x^{\mathsf{lt}(c_j)}) \enspace .$$

Here, the right hand side $\nabla f_j(x^{\mathsf{lt}(c_j)})$ has only an outdatedness of $t-\mathsf{lt}(c_j)$ compared with $\nabla f_j(x^{t-1})$, and $t-\mathsf{lt}(c_j)$ can be as small as $s$ if all the $s$ clusters are of equal size. In sum, this smaller outdatedness implies that our choice of $\widetilde{\nabla}^{t-1}$ is a better estimator of the full gradient.

Compared with Option I which updates $\zeta_c$ incrementally, our Option II can be seen as a batch-update version. An almost identical argument from the one above implies $\nabla f_j(\widetilde{x}) + \zeta_{c_j}$ in Option II is a guess of $\nabla f_j(x^{t-1})$ with outdatedness at most $s$, because $\overline{x}$ and $x^{t-1}$ differ by at most $s$ iterations. This is simpler to analyze because in Option I, the outdatedness is at most $s$ only when the clusters are of equal size. For the sake of cleanness, in this paper we only provide theoretical proofs for Option II. From an efficiency perspective, Option I is preferred in practice because it only needs one stochastic gradient computation per iteration where Option II needs two in the amortized sense.

### B.3 Convergence Analysis for ClusterSVRG

We provide theoretical upper bounds on the gradient estimator of ClusterSVRG, and compare it with SVRG. We make the following assumption in this section:

**Assumption B.3.** We assume that in each epoch of ClusterSVRG, between two consecutive iterations $t-1$ and $t$, it always satisfies $\|x^{t-1} - x^t\|^2 \le \xi$.

Note that this is a reasonable assumption because in most interesting applications the gradients are bounded. Under this assumption, we have the following measurement of the outdatedness:

**Claim B.4** (Outdatedness)**.**

*(a) For every $i \in [n]$, it satisfies $\|\nabla f_i(\overline{x}) - \nabla f_i(x^{t-1})\|^2 \le L^2 s^2 \xi$.*

*(b) Given any $(s,\sigma)$ raw clustering of the functions, for indices $i$ and $j$ that are randomly selected from the same cluster $S_c$, it satisfies $\mathbb{E}_{i,j}\big[\|\nabla f_i(\overline{x}) - \nabla f_j(x^{t-1})\|^2\big] \le 2\sigma + 2L^2 s^2 \xi$.*

The above claim captures the "outdatedness" parameter that we discussed before. Indeed, since $\overline{x}$ and $x^{t-1}$ can be at most $s$ iterations apart, the gradient $\nabla f_i(\overline{x})$ is outdated from $\nabla f_i(x^{t-1})$ by at most $s$ iterations. However, knowing only the number of iterations it remains difficult to measure the actual distance $\|\nabla f_i(\overline{x}) - \nabla f_i(x^{t-1})\|^2$. In Claim B.4.a, we simply use the smoothness of the function $f_i(\cdot)$ to provide a very loose upper bound on this distance, which turns out to be already quite useful when providing a theoretical comparison between ClusterSVRG and SVRG later in this section. In addition, Claim B.4.b simply combines Claim B.4.a with the $\sigma$-rawness of the given clustering. A formal proof is as follows.

*Proof of Claim B.4.* The first half of the claim follows from

$$\|\nabla f_i(\overline{x}) - \nabla f_i(x^{t-1})\|^2 \le L^2 \|\overline{x} - x^{t-1}\|^2 \le L^2 s^2 \xi^2 \ .$$

Above, the first inequality is due to the smoothness of function $f_i(\cdot)$, and the second inequality is because there are at most $s$ iterations between $\overline{x}$ and $x^{t-1}$.

To prove the second half, we compute that

$$\begin{aligned}
\mathbb{E}[\|\nabla f_i(\overline{x}) - \nabla f_j(x^{t-1})\|^2] &= \mathbb{E}[\|\nabla f_i(\overline{x}) - \nabla f_j(\overline{x}) + \nabla f_j(\overline{x}) - \nabla f_j(x^{t-1})\|^2] \\
&\le 2\mathbb{E}[\|\nabla f_i(\overline{x}) - \nabla f_j(\overline{x})\|^2] + 2\mathbb{E}[\|\nabla f_j(\overline{x}) - \nabla f_j(x^{t-1})\|^2] \\
&\le 2\sigma + 2L^2 s^2 \xi \ . 
\end{aligned}$$

□

We are now ready to state the main lemma of this section. It requires a few careful applications of Claim B.4, and we defer its proof to Appendix C.

**Lemma B.5** (Variance of ClusterSVRG)**.** *In ClusterSVRG with Option II, if a $(s,\sigma)$ raw clustering on the functions is provided and Assumption B.3 is satisfied, then we have the following upper bound on the variance of the gradient estimator at iteration $t$:*

$$\mathbb{E}[\mathbf{Var}_i[\widetilde{\nabla}^{t-1}]] = \mathbb{E}\big[\big\|\widetilde{\nabla}^{t-1} - \nabla f(x^{t-1})\big\|^2\big] \leq O\big(\sigma + L^2 s^2 \xi\big) \ . \tag{UB1}$$

*Where the expectation is over all the randomness in the current epoch. In addition, we also have*

$$\mathbb{E}[\mathbf{Var}_i[\widetilde{\nabla}^{t-1}]] \leq O(L) \cdot \mathbb{E}\Big[\big(f(\widetilde{x}) - f(x^*)\big) + \big(f(\overline{x}) - f(x^*)\big) + \big(f(x^{t-1}) - f(x^*)\big)\Big] \ . \tag{UB2}$$

We compare our main lemma to its counterpart for SVRG:

**Lemma B.6** (Variance of SVRG)**.** *In SVRG, we have the following upper bound on the variance of the gradient estimator at iteration $t$:*

$$\mathbb{E}[\mathbf{Var}_i[\widetilde{\nabla}^{t-1}]] = \mathbb{E}\big[\big\|\widetilde{\nabla}^{t-1} - \nabla f(x^{t-1})\big\|^2\big] \leq O\big(L^2 m^2 \xi\big) \ . \tag{UB1'}$$

*In addition, we also have*

$$\mathbb{E}[\mathbf{Var}_i[\widetilde{\nabla}^{t-1}]] \leq O(L) \cdot \mathbb{E}\big[\big(f(x^{t-1}) - f(x^*)\big) + \big(f(\widetilde{x}) - f(x^*)\big)\big] \ . \tag{UB2'}$$

*Proof.* The proof for (UB2') can be found on page 6 of [13]. The proof for (UB1') is almost identical to that for (UB1). □

Following the main intuition behind all the variance-reduction methods, we know that a stochastic gradient method converges faster if the variance of the gradient estimator $\widetilde{\nabla}^{t-1}$ is small. It is easy to see that (UB1) is a better upper bound than (UB1') if an $(s,\sigma)$ raw clustering is provided, and at the same time our (UB2) has the same form as (UB2'). Therefore, the variance of the gradient estimator in ClusterSVRG is better than that in SVRG.

For interested readers, we compare these upper bounds more carefully in the next subsection.

## B.4 Discussion

Upper bounds (UB2) and (UB2') are smaller than (UB1) and (UB1') respectively for large iterations $t$, because our objective value $f(x)$ approaches to $f(x^*)$ as $t$ increases. For this reason, they are better choices in order to obtain an asymptotic worst-case running time.

For instance, if *only* (UB2') is used in the analysis, Johnson and Zhang [13] showed that for $\sigma$-strongly convex objectives $F(x)$ in (4.1), by setting $m = L/\sigma$, the SVRG method converges to an $\varepsilon$-approximate minimizer of $F(x)$ in $T = O((n + \frac{L}{\sigma})\log(1/\varepsilon))$ iterations. This matches the best known running time performance for non-accelerated methods on solving (4.1). For ClusterSVRG, using (UB2) and an analogous analysis to [13], it matches the worst case running time of SVRG.

In contrast, in the first a few epochs of the algorithm, we expect (UB1) and (UB1') to be smaller than (UB2) and (UB2') respectively, thus providing tighter upper bounds on the variance. For this reason, if $\sigma$ is relatively small, because $L^2 s^2 \xi$ is significantly smaller than $L^2 m^2 \xi$, we expect ClusterSVRG to have a much smaller variance than SVRG in the first a few epochs, thus outperform SVRG. Such outperforance is indeed observed in our experiment in Section 5.

**Interpolation Between SVRG and SAGA.** When every index $i$ is itself a cluster $S_i = \{i\}$, and if we choose $m = \infty$ so there is only one epoch, ClusterSVRG with Option I reduces to SAGA, because $\nabla f_j(\widetilde{x})+\zeta_{c_j}$ becomes exactly $\nabla f_j(\mathsf{lt}(j))$ where $\mathsf{lt}(j)$ is the last time index $j$ was chosen. In this case, it is not necessary to compute snapshots anymore. Also, recall that if there is only one cluster $S_1=[n]$, our ClusterSVRG is identical to SVRG by comparing (B.2) with (B.1).

In sum, SAGA and SVRG are two extreme variants of ClusterSVRG. Our experiments in Section 5 confirm that SVRG outperforms SAGA when the data points are well clustered and vice versa.

# C Proof of Lemma B.5

**First half of Lemma B.5.** We first prove inequality (UB1), the first half of Lemma B.5. Suppose that $i \in S_c$ so the random index $i$ belongs to the $c$-th cluster. In this case, we can rewrite

$$\begin{aligned}
&\widetilde{\nabla}^{t-1} - \nabla f(x^{t-1}) \\
&= \frac{1}{n}\sum_{j=1}^{n}\big(\nabla f_j(\widetilde{x}) + \zeta_{c_j}\big) + \nabla f_i(x^{t-1}) - \nabla f_i(\widetilde{x}) - \zeta_{c_i} - \nabla f(x^{t-1}) \\
&= \frac{1}{n}\sum_{j\in[n]\setminus S_c}\Big(\nabla f_j(\widetilde{x}) + \zeta_{c_j} - \nabla f_j(x^{t-1})\Big) \\
&\quad + \frac{1}{n}\sum_{j\in S_c}\big(\nabla f_j(\widetilde{x}) + \zeta_{c_j} - \nabla f_j(x^{t-1})\big) + \nabla f_i(x^{t-1}) - \nabla f_i(\widetilde{x}) - \zeta_{c_i} \\
&= \frac{1}{n}\sum_{c'\neq c}\sum_{j\in S_{c'}}\Big(\nabla f_j(\widetilde{x}) + \zeta_{c'} - \nabla f_j(x^{t-1})\Big) + \frac{n-n_c}{n}\big(\nabla f_i(\widetilde{x}) - \zeta_c - f_i(x^{t-1})\big) \\
&\quad + \frac{1}{n}\sum_{j\in S_c}\big(\nabla f_j(\widetilde{x}) - \nabla f_i(\widetilde{x})\big) + \frac{1}{n}\sum_{j\in S_c}\big(\nabla f_i(x^{t-1}) - \nabla f_j(x^{t-1})\big) \enspace .
\end{aligned}$$

Therefore, conditioning on that $i$ belongs to $S_c$, we can upper bound the variance as follows:

$$\begin{aligned}
&\mathbb{E}\bigg[\Big\|\widetilde{\nabla}^{t-1} - \nabla f(x^{t-1})\Big\|^2 \,\bigg|\, i\in S_c\bigg] \\
&\leq \underbrace{4\,\mathbb{E}\bigg[\Big\|\frac{1}{n}\sum_{c'\neq c}\sum_{j\in S_{c'}}\Big(\nabla f_j(\widetilde{x}) + \zeta_{c'} - \nabla f_j(x^{t-1})\Big)\Big\|^2\bigg]}_{(\spadesuit)} + \underbrace{4\,\mathbb{E}\bigg[\Big\|\frac{n-n_c}{n}\big(\nabla f_i(\widetilde{x}) - \zeta_c - f_i(x^{t-1})\big)\Big\|^2 \,\bigg|\, i\in S_c\bigg]}_{(\heartsuit)} \\
&\quad + \underbrace{4\,\mathbb{E}\bigg[\Big\|\frac{1}{n}\sum_{j\in S_c}\big(\nabla f_j(\widetilde{x}) - \nabla f_i(\widetilde{x})\big)\Big\|^2 \,\bigg|\, i\in S_c\bigg]}_{(\clubsuit)} + \underbrace{4\,\mathbb{E}\bigg[\Big\|\frac{1}{n}\sum_{j\in S_c}\big(\nabla f_i(x^{t-1}) - \nabla f_j(x^{t-1})\big)\Big\|^2 \,\bigg|\, i\in S_c\bigg]}_{(\diamondsuit)} \enspace .
\end{aligned} \tag{C.1}$$

Above, the inequality is because for every $q$ vectors $v_1,\dots,v_q$, it satisfies

$$\Big\|\sum_{k=1}^{q} v_k\Big\|^2 \leq q\sum_{k=1}^{q}\|v_k\|^2 \enspace . \tag{C.2}$$

Now, the definition of $\sigma$ raw clustering on functions immediately imply that

$$(\clubsuit) = \mathbb{E}\bigg[\Big\|\frac{1}{n}\sum_{j\in S_c}\big(\nabla f_j(\widetilde{x}) - \nabla f_i(\widetilde{x})\big)\Big\|^2 \,\bigg|\, i\in S_c\bigg] \leq \frac{n_c^2}{n^2}\mathbb{E}_{i,j\in_R S_c}\big[\|\nabla f_i(\widetilde{x}) - \nabla f_j(\widetilde{x})\|^2\big] \leq \frac{n_c^2}{n^2}\sigma \enspace .$$

Above, the first inequality follows from (C.2). Similarly, we also have

$$(\diamondsuit) = \mathbb{E}\Big[\Big\|\frac{1}{n}\sum_{j\in S_c}\big(\nabla f_i(x^{t-1}) - \nabla f_j(x^{t-1})\big)\Big\|^2\Big| i \in S_c\Big] \le \frac{n_c^2}{n^2}\sigma \enspace .$$

To upper bound the other two terms in (C.1), we first compute that

$$\begin{aligned}
(\heartsuit) &= \frac{(n-n_c)^2}{n^2}\mathbb{E}_{i,k\in_R S_c}\Big[\big\|\big(\nabla f_i(\widetilde{x}) - \nabla f_k(\widetilde{x})\big) - \big(\nabla f_i(x^{t-1}) - \nabla f_k(\overline{x})\big)\big\|^2\Big] \\
&\le \frac{2(n-n_c)^2}{n^2}\Big(\mathbb{E}_{i,k\in_R S_c}\Big[\big\|\nabla f_i(\widetilde{x}) - \nabla f_k(\widetilde{x})\big\|^2\Big] + \mathbb{E}_{i,k\in_R S_c}\Big[\big\|\nabla f_i(x^{t-1}) - \nabla f_k(\overline{x})\big\|^2\Big]\Big) \\
&\le O\big(\sigma + L^2 s^2 \xi\big) \enspace .
\end{aligned}$$

Here, the last inequality uses Claim B.4. Similarly, we can compute that

$$\begin{aligned}
(\spadesuit) &\overset{①}{\le} \frac{n-n_c}{n^2}\mathbb{E}\Big[\sum_{c'\neq c}\sum_{j\in S_{c'}}\Big\|\nabla f_j(\widetilde{x}) + \zeta_{c'} - \nabla f_j(x^{t-1})\Big\|^2\Big] \\
&= \frac{n-n_c}{n^2}\sum_{c'\neq c} n_{c'}\cdot\mathbb{E}_{j,k\in_R S_{c'}}\Big[\Big\|\big(\nabla f_j(\widetilde{x}) - \nabla f_k(\widetilde{x})\big) - \big(\nabla f_j(x^{t-1}) - \nabla f_k(\overline{x})\big)\Big\|^2\Big] \\
&\le \frac{2(n-n_c)}{n^2}\sum_{c'\neq c} n_{c'}\cdot\mathbb{E}_{j,k\in_R S_{c'}}\Big[\big\|\nabla f_j(\widetilde{x}) - \nabla f_k(\widetilde{x}))\big\|^2 + \big\|\nabla f_j(x^{t-1}) - \nabla f_k(\overline{x})\big\|^2\Big] \\
&\overset{②}{\le} O\big(\sigma + L^2 s^2 \xi\big) \enspace .
\end{aligned}$$

Above, ① uses inequality (C.2), and ② uses Claim B.4. Substituting all of the four upper bound back to inequality (C.1), we immediately have that

$$\mathbb{E}\left[\Big\|\widetilde{\nabla}^{t-1} - \nabla f(x^{t-1})\Big\|^2\Big| i \in S_c\right] \le O\big((\spadesuit) + (\heartsuit) + (\clubsuit) + (\diamondsuit)\big) = O\big(\sigma + 2L^2 s^2 \xi\big) \enspace .$$

Taking expectation over all possible $c = 1, 2, \dots, s$, we get our desired upper bound.
**Second half of Lemma B.5.** We now prove inequality (UB2), the second half of Lemma B.5. This time, we rewrite

$$\begin{aligned}
&\widetilde{\nabla}^{t-1} - \nabla f(x^{t-1}) \\
&= \frac{1}{n}\sum_{j=1}^{n}\big(\nabla f_j(\widetilde{x}) + \zeta_{c_j}\big) + \nabla f_i(x^{t-1}) - \nabla f_i(\widetilde{x}) - \zeta_{c_i} - \nabla f(x^{t-1}) \\
&= \big(f(\widetilde{x}) - f(x^{t-1})\big) + \big(\nabla f_i(x^{t-1}) - \nabla f_i(\widetilde{x})\big) - \zeta_{c_i} + \frac{1}{n}\sum_{c'=1}^{s} n_{c'}\zeta_{c'} \enspace ,
\end{aligned}$$

and therefore

$$\begin{aligned}
&\mathbb{E}\Big[\big\|\widetilde{\nabla}^{t-1} - \nabla f(x^{t-1})\big\|^2\Big] \\
&\le 4\mathbb{E}\Big[\big\|\nabla f(\widetilde{x}) - \nabla f(x^{t-1})\big\|^2 + \big\|\nabla f_i(x^{t-1}) - \nabla f_i(\widetilde{x})\big\|^2 + \|\zeta_{c_i}\|^2 + \Big\|\frac{1}{n}\sum_{c'=1}^{s} n_{c'}\zeta_{c'}\Big\|^2\Big] \enspace . \qquad \text{(C.3)}
\end{aligned}$$

We next upper bound the four terms on the right hand side of (C.3). The first two terms enjoy the following simple upper bounds due to the SVRG paper [13]:

$$\begin{aligned}\big\|\nabla f(\widetilde{x}) - \nabla f(x^{t-1})\big\|^2 &\overset{①}{\leq} 2\big\|\nabla f(\widetilde{x}) - \nabla f(x^*)\big\|^2 + 2\big\|\nabla f(x^{t-1}) - \nabla f(x^*)\big\|^2 \\ &\overset{②}{\leq} 4L\big(f(\widetilde{x}) - f(x^*)\big) + 4L\big(f(x^{t-1}) - f(x^*)\big) \enspace.\end{aligned}$$

$$\begin{aligned}&\mathbb{E}\big[\big\|\nabla f_i(x^{t-1}) - \nabla f_i(\widetilde{x})\big\|^2\big] \\ &\overset{③}{\leq} 2\mathbb{E}\big[\big\|\nabla f_i(x^{t-1}) - \nabla f_i(x^*)\big\|^2\big] + 2\mathbb{E}\big[\big\|\nabla f_i(\widetilde{x}) - \nabla f_i(x^*)\big\|^2\big] \\ &\overset{④}{\leq} 2\mathbb{E}\big[f_i(x^{t-1}) - f_i(x^*) - \langle f_i(x^*), x^{t-1} - x^*\rangle\big] + 2\mathbb{E}\big[f_i(\widetilde{x}) - f_i(x^*) - \langle f_i(x^*), \widetilde{x} - x^*\rangle\big\|^2\big] \\ &= 4L \cdot \mathbb{E}\big[\big(f(x^{t-1}) - f(x^*)\big) + \big(f(\widetilde{x}) - f(x^*)\big)\big] \enspace.\end{aligned}$$

Above, ① and ③ follow from (C.2), while ② and ④ follow from a classical inequality for smooth functions.[15]

In fact, the other two terms can be bounded using similar ideas from the ones above. For instance,

$$\begin{aligned}&\mathbb{E}\big[\big\|\zeta_{c_i}\big\|^2\big] = \mathbb{E}\big[\|\nabla f_k(\overline{x}) - \nabla f_k(\widetilde{x})\|^2 \,\big|\, k \text{ is randomly chosen in } [n]\big] \\ &\leq 2\mathbb{E}\big[\big\|\nabla f_k(\overline{x}) - \nabla f_k(x^*)\big\|^2\big] + 2\mathbb{E}\big[\big\|\nabla f_k(\widetilde{x}) - \nabla f_k(x^*)\big\|^2\big] \\ &\leq 2\mathbb{E}\big[f_k(\overline{x}) - f_k(x^*) - \langle f_k(x^*), \overline{x} - x^*\rangle\big] + 2\mathbb{E}\big[f_k(\widetilde{x}) - f_k(x^*) - \langle f_k(x^*), \widetilde{x} - x^*\rangle\big\|^2\big] \\ &= 4L \cdot \mathbb{E}\big[\big(f(\overline{x}) - f(x^*)\big) + \big(f(\widetilde{x}) - f(x^*)\big)\big] \enspace.\end{aligned}$$

One can similarly prove that

$$\mathbb{E}\Big[\Big\|\frac{1}{n}\sum_{c'=1}^{s} n_{c'}\zeta_{c'}\Big\|^2\Big] \leq O(L) \cdot \mathbb{E}\big[\big(f(\overline{x}) - f(x^*)\big) + \big(f(\widetilde{x}) - f(x^*)\big)\big] \enspace.$$

Substituting these upper bounds back to (C.3) concludes the proof. □

[15] See for instance in Theorem 2.1.5 of the textbook [20]: if $g$ is convex and $L$-smooth, with minimizer $x^*$, then for every $x, y$ we we have $\|\nabla g(x) - \nabla g(y)\|^2 \leq 2L(g(x) - g(y) - \langle\nabla g(y), x - y\rangle)$.

| dataset | Covtype | | | SensIT | | | News20 | | |
|---|---|---|---|---|---|---|---|---|---|
| $n$ | 581,012 | | | 78,823 | | | 19,996 | | |
| $d$ | 54 | | | 100 | | | 1,355,191 | | |
| sparsity | 21.60% | | | 99.01% | | | 0.03% | | |
| one-pass time of SAGA | 0.49s | | | 0.26s | | | 1.06s | | |
| radius $\delta$ | 0.02 | 0.06 | 0.17 | 0.1 | 0.2 | 0.4 | 0.5 | 0.7 | 0.9 |
| detection result | yes | yes | yes | no | yes | yes | no | no | no[16] |
| clustering computation time | 1.55<br>3.15 | 1.53<br>3.12 | 0.84<br>(sec)<br>1.72<br>(pass) | -<br>- | 0.86<br>3.32 | 0.30<br>(sec)<br>1.14<br>(pass) | -<br>- | -<br>- | -<br>- |

Table 1: Time needed for raw clustering. We use the one-pass running time of SAGA as benchmark in order to take into account different data sizes. We emphasize that the clustering time shall become even more negligible, after being amortized over multiple runs due to different analysis tasks, parameter tunings, etc.

| dataset | Covtype | | | SensIT | | | News20 | | |
|---|---|---|---|---|---|---|---|---|---|
| radius $\delta$ | 0.02 | 0.06 | 0.17 | 0.1 | 0.2 | 0.4 | 0.5 | 0.7 | 0.9 |
| Haar transformation time | 0.66s | 0.91s | 1.25s | 0.61s | 0.78s | 0.89s | 0.64s | 0.73s | 1.12s |
| Haar time / one-pass | 1.34 | 1.85 | 2.55 | 2.34 | 3.00 | 3.42 | 0.60 | 0.69 | 1.06 |

Table 2: Time needed for Haar transformation in ClusterACDM. We emphasize that the transformation time shall become even more negligible, after being amortized over multiple runs due to different analysis tasks, parameter tunings, etc.

**Algorithm 3** Compute a Raw Clustering

**Input:** $n$ data samples $a_1, \cdots, a_n$.

```
DB ← {}.
for i ← 1 to n do
    a_j ← FindNearestNeighbor(a_i, DB, R)
    if a_j exists then
        add a_i into the cluster of a_j
    else
        add a_i into DB as a new cluster
    end if
end for
```

# D Our Particular Raw-Clustering Algorithm

Let FindNearestNeighbor$(v, DB, R)$ be an oracle that outputs either a close neighbor of $v$ with distance at most $R$ in the set $DB$ or nothing. We need to ensure that the oracle is able to find a close neighbor with high probability if such neighbor exists. There are many fast nearest neighbor algorithms satisfying this property, such as LSH [2, 3] and product quantization [10, 12]. In our experiments, we use E2LSH [2] with sparse vector supports.

Given $n$ data vectors $a_1, \dots, a_n$, we iteratively call FindNearestNeighbor$(a_i, DB, R)$ for each

[16] Recall that News20 is included in our experiments only for comparison purpose, and simply using $0.3T$ time, the detecting phase reports that no good clustering structure has been found on News20. For such datasets, it is not worth computing clustering in full because Algorithm 3 can keep inserting new clusters which costs the running time to blow up. However, as shown in Figure 6 and 7, having such clustering information never hurts the running time of ClusterSVRG or ClusterACDM.

$i = 1, 2, \ldots, n$. If FindNearestNeighbor outputs a neighbor, we add $a_i$ into the corresponding cluster of that neighbor; otherwise we create a new cluster that only contains $a_i$. See Algorithm 3. We emphasize that since we only need raw clustering in this paper, it is perfectly fine if the oracle misses some true neighbors. As a consequence, *approximate* neighbor finding algorithms such as LSH perfectly suit our purposes.

We can use Algorithm 3 for both the detection phase and the finding phase.

- Detecting Phase. Given input $\delta > 0$, we run Algorithm 3 on randomly sampled data vectors without replacement, for a limited time $0.3T$, where $T$ is the one-pass running time of SAGA. We obtain the number of clusters $s'$, and the number of total sampled points $n'$. If $s'/n' \leq 10\%$, we know that the dataset has a good clustering structure and we continue to the computing phase.

- Computing Phase. Given input $\delta > 0$, we run Algorithm 3 on the entire dataset and output the clusters.

From Table 1, we confirm that after the dataset is known to have good clustering structure (such as Covtype and SensIT), only a total of around three-pass running time of SAGA is needed to compute the entire clustering. Here we picked the best parameters for running LSH in each case.